\documentclass[nohyperref]{article}

\usepackage{microtype}
\usepackage{graphicx}
\usepackage{subfigure}
\usepackage{booktabs} 
\usepackage{tabularx} 
\usepackage[table]{xcolor}
\usepackage{multirow}
\usepackage{makecell}
\usepackage{hyperref}
\usepackage{booktabs}
\usepackage{bbding}
\usepackage{float}
\usepackage{listings}

\usepackage[accepted]{icml2023}

\usepackage{amsmath}
\usepackage{amssymb}
\usepackage{mathtools}
\usepackage{amsthm}

\usepackage[capitalize,noabbrev]{cleveref}
\newcommand{\papername}[0]{Measuring The Impact Of Programming Language Distribution}
\theoremstyle{plain}

\theoremstyle{definition}

\theoremstyle{remark}

\usepackage[textsize=tiny]{todonotes}

\newcommand{\babelcode}{BabelCode}
\definecolor{codegreen}{rgb}{0,0.6,0}
\definecolor{codegray}{rgb}{0.5,0.5,0.5}
\definecolor{codepurple}{rgb}{0.58,0,0.82}
\definecolor{backcolour}{rgb}{0.95,0.95,0.92}
\definecolor{greencheck}{HTML}{008009}
\lstdefinestyle{mystyle}{
    backgroundcolor=\color{backcolour},   
    commentstyle=\color{codegreen},
    keywordstyle=\color{magenta},
    numberstyle=\tiny\color{codegray},
    stringstyle=\color{codepurple},
    basicstyle=\ttfamily\footnotesize,
    breakatwhitespace=false,         
    breaklines=true,                 
    captionpos=b,                    
    keepspaces=true,                 
    numbers=left,                    
    numbersep=5pt,                  
    showspaces=false,                
    showstringspaces=false,
    escapechar={|}, 
    showtabs=false,                  
    tabsize=2
}
\newcommand{\RedX}[0]{{\color{red} \XSolid}}
\newcommand{\GreenCheck}[0]{{\color{greencheck} \Checkmark}}

\icmltitlerunning{ \hfill \papername \hfill \thepage}

\begin{document}

\twocolumn[
\icmltitle{\papername}



\icmlsetsymbol{atgoogle}{*}
\icmlsetsymbol{}{}

\begin{icmlauthorlist}
\icmlauthor{Gabriel Orlanski}{nyu,google-lab,atgoogle}
\icmlauthor{Kefan Xiao}{google-lab}
\icmlauthor{Xavier Garcia}{google-brain}
\icmlauthor{Jeffrey Hui}{google-lab}
\icmlauthor{Joshua Howland}{google-lab}
\icmlauthor{Jonathan Malmaud}{google-lab}
\icmlauthor{Jacob Austin}{google-brain}
\icmlauthor{Rishabh Singh}{google-lab,atgoogle}
\icmlauthor{Michele Catasta}{google-lab,atgoogle}
\end{icmlauthorlist}

\icmlaffiliation{nyu}{Department of Computer Science, New York University, New York, New York}
\icmlaffiliation{google-lab}{Google Labs}
\icmlaffiliation{google-brain}{Google Brain}

\icmlcorrespondingauthor{Gabriel Orlanski}{go533@nyu.edu}
\icmlcorrespondingauthor{Kefan Xiao}{kfxiao@google.com}
\icmlcorrespondingauthor{Xavier Garcia}{xgarcia@google.com}

\icmlkeywords{Machine Learning, ICML}

\vskip 0.3in
]



\printAffiliationsAndNotice{*Work Done While At Google} 

\begin{abstract}
Current benchmarks for evaluating neural code models focus on only a small subset of programming languages, excluding many popular languages such as Go or Rust. To ameliorate this issue, we present the \babelcode{} framework for execution-based evaluation of any benchmark in any language. \babelcode{} enables new investigations into the qualitative performance of models' memory, runtime, and individual test case results. Additionally, we present a new code translation dataset called Translating Python Programming Puzzles (TP3) from the Python Programming Puzzles \citep{schuster2021programming} benchmark that involves translating expert-level python functions to any language. With both \babelcode{} and the TP3 benchmark, we investigate if balancing the distributions of 14 languages in a training dataset improves a large language model's performance on low-resource languages. Training a model on a balanced corpus results in, on average, 12.34\% higher $pass@k$ across all tasks and languages compared to the baseline. We find that this strategy achieves 66.48\% better $pass@k$ on low-resource languages at the cost of only a 12.94\% decrease to high-resource languages. In our three translation tasks, this strategy yields, on average, 30.77\% better low-resource $pass@k$ while having 19.58\% worse high-resource $pass@k$.\footnote{\href{https://github.com/google-research/babelcode}{https://github.com/google-research/babelcode}}
\end{abstract}

\section{Introduction}
\begin{figure*}[h]
    \centering
    \caption{Overview of this work's contributions. }
    \includegraphics[width=\linewidth]{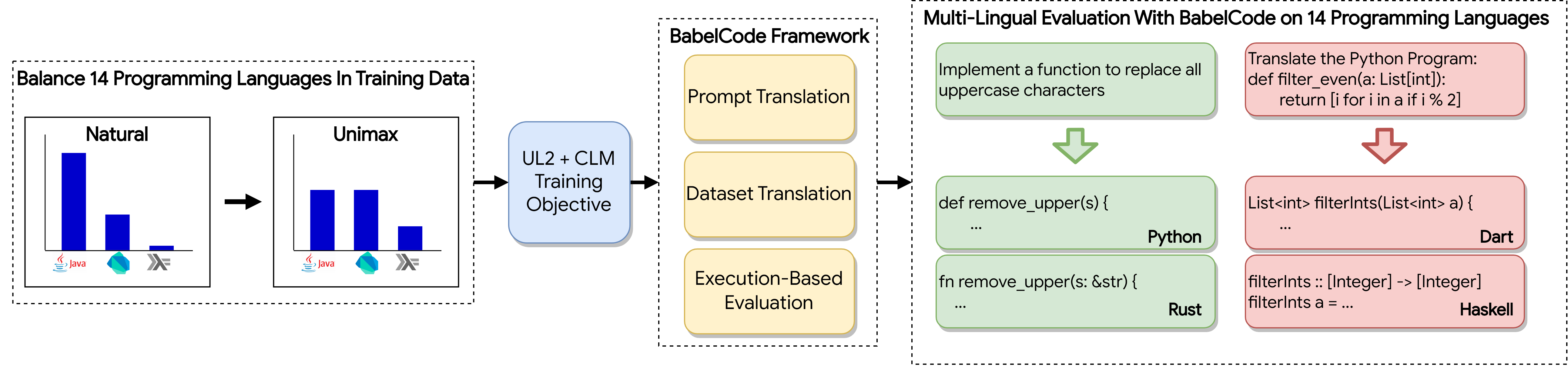}
    \label{fig:overview}
\end{figure*}

In the \href{https://insights.stackoverflow.com/survey}{2022 StackOverflow Developer Survey}, Rust was the 14th most popular programming language despite not ranking in the survey taken five years prior. However, the 13th most popular language, Go, has nearly doubled Rust's number of StackOverflow questions in this time frame. Further, despite their similar popularity, Go has nearly 350\% more source code available \citep{kocetkov2022stack}. These disparities highlight the problem that many popular programming languages are starkly low-resource, especially compared to the most popular languages. 

Despite their impressive generative capabilities, especially in code, Large Language Models (LLM) are adversely impacted by this language resource imbalance. Thus, developers will likely find minimal utility from LLMs if they are not using the extremely popular languages. It is therefore imperative to investigate how to mitigate the discrepancy between a language's popularity and the amount of data available for it. Prior works focusing on code generation \citep{ahmad2021unified} and multilingual natural language processing \citep{arivazhagan2019massively,Conneau2019UnsupervisedCR} use temperature-based strategies to balance the training languages. Such a strategy duplicates extremely low-resource languages thousands of times, which has been shown to significantly reduce performance \citep{allamanis2019adverse}. 

Beyond the the language balancing strategy, evaluating code LLMs in a multi-lingual setting presents significant challenges. Existing datasets are either mono-lingual \citep{Chen2021EvaluatingLL,austin2021program,Lai2022DS1000AN} or limited to only a subset of popular programming languages \citep{roziere2020unsupervised}. Each problem in these datasets, which we henceforth refer to as a \emph{benchmark}, contains an input, and a canonical solution along with the test-cases for checking correctness. Creating a new benchmark for each language of interest would require insurmountable engineering and monetary costs. To address both of these problems, we present the \babelcode{} framework for execution-based evaluation of \textit{any benchmark} in \textit{any language} and use it to investigate the impact of programming language distribution on code generation and translation. 

\babelcode{} is open-sourced, has an extensive test suite, and supports evaluating four benchmarks in 14 languages. It is designed specifically to enable future research directions such as the evaluation of custom data-structures. \babelcode{} allows investigation of novel research directions through the measurement of memory and runtime usage for a given prediction, as well as the outcomes of individual test cases. Furthermore, we can use \babelcode{} to build multi-lingual execution based benchmarks from existing mono-lingual datasets. We demonstrate this functionality by creating a new dataset called Translating Python Programming Puzzles (TP3) from the Python Programming Puzzles \citep{schuster2021programming} benchmark, where the objective is to translate expert-level python programs to other languages. The source programs for TP3 are the hand-crafted verification functions for each problem in P3. As the authors hand-wrote each function, they are significantly more complex than the current state-of-the-art code translation benchmarks, such as Transcoder \citep{roziere2020unsupervised}, for which code LLMs are already achieving highly impressive results.

Our presented framework is closely related to the concurrent work of MBXP \citep{athiwaratkun2023multilingual} and Multi-PLE\citep{cassano2022scalable}. While MBXP is quite similar to \babelcode{}, it is not open-sourced and requires that the input benchmarks be in Python. Multi-PLE is open-sourced, but only supports generation tasks and contains significant errors in multiple languages. \babelcode{} addresses these issues through an extensive test suite that ensures that the code generated is correct, and that crucial functionality, such as data structure equivalence, works when executed.

With the \babelcode{} framework, we investigate remedies to the problems of programming language imbalance. We utilize the Unimax algorithm \citep{chung2023unimax} to limit the maximum number of times to duplicate a language's data to a constant $N$. We then train 1B, 2B, and 4B parameter decoder-only models on both the natural and Unimax $N$ distributions. We utilize the UL2 \citep{tay2022unifying} and causal language modeling training objective. We find that models trained on the balanced dataset significantly outperform the baseline models on low-resource languages across all tasks. Further, we find that the resulting performance drop on high-resource languages is mitigated by increasing the model size.

This paper makes the following key contributions:
\begin{itemize}
    \item We propose and release \babelcode{}, a new execution-based evaluation framework that allows for multilingual evaluation of code generation and translation capabilities of code language models. It also supports the easy addition of new benchmark tasks and execution-based metrics.
    \item We show that the code language models trained on the natural distributions of GitHub source code have poor performance on low-resource languages in both generation and translation tasks.
    \item We propose a new data balancing strategy for programming languages to improve performance on low-resource languages. We demonstrate that the resulting models outperform the baseline models across all tasks by an average of 12.34\% $pass@k$ for all languages, with a further improvement of 39.70\% $pass@k$ to low-resource languages.
    \item We find that the average improvements on low-resource languages from training on balanced data do not scale with model size. But scaling model sizes significantly helps the average $pass@k$ loss compared to the baselines on high-resource languages going from a loss of 39.70\% with the 1B model to a loss of 2.47\% with the 4B model.
\end{itemize}

\section{The \babelcode{} Framework}
\begin{figure}[h]
    \centering
    \caption{\babelcode{}'s domain specific language for representing the input and output types of a question. Prior works require that the source dataset be written in Python, while our DSL removes this restriction and allows users to create datasets in \textit{any} language. This enables seamless additions of new languages while simplifying future expansions to features such as custom data structures.}
    \label{fig:dsl}
    \includegraphics[width=0.48\textwidth]{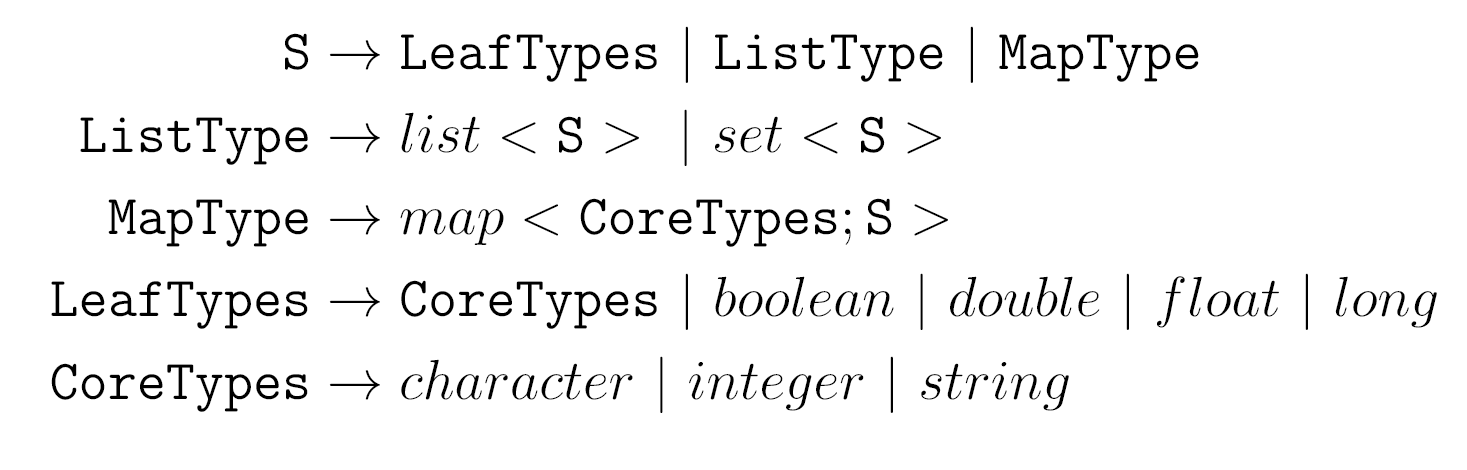}
\end{figure}
\begin{table*}[h]
    \centering 
    \caption{Differences between \babelcode{} and prior works. NL2C is natural language to code, while C2C is code to code datasets. \babelcode{} has an extensive test-suite that automatically tests each language's implementation and correctness when executed.}
    \begin{tabular}{l|cccccccc}
        \toprule
        & Open& \#  & NL2C & C2C & Mem. \& & Test & Indiv. Test & Lang. Agnostic\\
        Name & Sourced & Lang. & Support & Support & Time Metrics & Suite  & Case Results & Datasets \\
        \midrule
        MultiPL-E & \GreenCheck & 18 & \GreenCheck  & \RedX & \RedX & \RedX & \RedX & \RedX\\
        MBXP & \RedX & 10 & \GreenCheck & \GreenCheck & \RedX & \RedX & \GreenCheck & \RedX \\
        \babelcode{} & \GreenCheck & 14 & \GreenCheck & \GreenCheck &  \GreenCheck & \GreenCheck & \GreenCheck & \GreenCheck
    \end{tabular}
    \label{tab:prior-works}
\end{table*}

\babelcode{} enables the evaluation of a collection of problems, each consisting of a prompt and a set of test cases, in any language through four stages: 1) represent each test case in our domain specific language (DSL) defined in \autoref{fig:dsl}, 2) use this generic form to generate the test cases in the target language from the input and output values, 3) use a Jinja\footnote{\href{https://jinja.palletsprojects.com/en/3.1.x/}{https://jinja.palletsprojects.com/en/3.1.x/}} template to generate a testing script in the target language, and 4) execute the target script through the command line. This is done autonomously, requiring minimal human intervention. We provide an overview of how an example problem is translated in \autoref{fig:bc-sample}. Overall the key novel elements of \babelcode{} are:  I) the use of a DSL to translate programming questions, II) type-specific equivalence, III) the ability to measure the performance of a given program at a low level (i.e., memory used, runtime, which tests passed), and IV) a large scale test-suite for ensuring correctness of generated code.

\subsection{Framework Design}\label{sec:bc-methods-trans}
\babelcode{} shares many design similarities to the concurrent work from \citet{athiwaratkun2023multilingual}. Specifically, we follow the same approach to inferring argument and return types. We follow the respective documentation and tutorials for each language to determine which native types to use. We also use these docs to determine the docstring formatting and naming convention. These mappings are used to generate unit and integration tests for each language automatically. They ensure that each language's implementation is syntactically correct and that, when executed, the equality comparison is correct. We describe the framework design and similarities to \citet{athiwaratkun2023multilingual} in \autoref{appendix:bc-design}.

\textbf{DSL Representations:} Using a DSL in the first phase, we do not force the inputs to be Python, thus enabling more flexibility to represent more generic tasks. For example, given the inputs from two test cases: {\tt \{"a":[[1],[],[80]]\}} and {\tt \{"a":[]\}}, we only represent the \textit{types} in our generic DSL. Thus, the resulting type string for this input is {\tt map<string;list<integer>>}. We do not represent the actual values in the generic form as we can easily translate literals across languages. This allows users to create a dataset from any language by requiring that they only represent the types of the inputs and outputs in this generic form. The language agnostic nature of the DSL enables future extensions of \babelcode{} to incorporate complex inputs and outputs such as custom data-structures. For example, the representation of a node class in a BST could be {\tt BSTNode<integer;integer>}.  

\textbf{Equality Checking:} We support floating point equivalence to a precision of $\epsilon=1\mathrm{e}{-6}$ for floats  and $\epsilon=1\mathrm{e}{-9}$ for doubles. To determine if a given value is a {\tt float} or a {\tt double}, we count the number of digits after the decimal place. We apply this same logic to {\tt int} and {\tt long} by counting the total number of digits. Languages such as C\# do not, by default, support deep equivalence of data structures. In such cases, we serialize the objects to JSON and check that the resulting strings are equal. Otherwise, we use the language built-in deep equality functionality. 

\textbf{Test Statement Execution:} We opt to print the result of each test case (i.e. {\tt TEST-0...PASSED}) to the standard output in a parseable format across all languages. Along with try-catch blocks, this allows the evaluation of \textit{every} test case for a given problem. This allows finer analysis of individual programs when compared to using {\tt assert} statements as it identifies if specific corner cases fail.

\textbf{Prompt Translation:} As \citet{Wang2022ReCodeRE} showed, LLMs are sensitive to the input prompts for code generation. Therefore \babelcode{} supports prompt translation and construction for multiple different problem formulations. We replace the names of languages, such as Python, with the target language. We use the language-specific naming convention to properly format the signature in the best practice style. If an argument uses a reserved keyword, we append {\tt arg} to its name so that it retains the same meaning but will no longer conflict. We replace Python-specific terms with their equivalent names in the target language.  For tasks formulated as code-completion, we support formatting the problem description as a native docstring. We do \textit{not} translate the {\tt import} statements in the header. Instead, we exclude the headers from all languages to provide a language-agnostic format. Translating prompts to a target language is not novel by itself, as both \citet{athiwaratkun2023multilingual} and \citet{cassano2022scalable} proposed methods to accomplish this. \babelcode{}'s builds on those works by translating reserved characters. For example, in Julia, the "{\tt \$}" in docstrings will raise errors if not properly escaped. Thus, we implement methods to automatically handle such cases and ensure correctness. 

\subsection{Differences To Prior Works}\label{sec:diff-prior}

We summarize the high-level differences between  \babelcode{} and prior works in \autoref{tab:prior-works}. The \textbf{MBXP} framework from \citet{athiwaratkun2023multilingual} is the most similar to our work as discussed in \autoref{sec:bc-methods-trans}. Similar to \babelcode{}, MBXP does have individual test-case results; however, it uses {\tt assert} statements and thus can only determine the first test-case that fails. MBXP does use language experts to review the generated code's quality and discuss the validation it supports to ensure that generated code parses and/or compiles for its respective language. \babelcode{} also has this functionality but, additionally, it ensures correctness through a test suite that covers the execution of generated code. We provide scripts to allow validating that source solutions to a dataset pass the generated code. For languages that do not have a solution in the respective dataset, we generate ``mock" predictions that return the expected output type. This allows us to ensure that generated code is correct in \textit{all} supported languages even if no solution exists.

The \textbf{MultiPL-E} framework from \citet{cassano2022scalable} supports 18 languages compared to \babelcode{}'s 16. However, we support four datasets,  while MultiPL-E only currently has support for two datasets. In addition, \babelcode{} also supports fine-grained evaluation metrics for memory, running time, and individual test cases. Our extensive test suite and validation scripts have also exposed many language-specific idiosyncrasies that naive methods of translation fail to handle. For example, in Julia, any ``{\tt \$}" will be treated as string interpolation, even if it is in a docstring. Thus, in the majority of cases, these must be escaped. We automatically rename variables that use reserved keywords. In languages such as C\#, the {\tt ==} operator checks equivalence by \textit{reference} instead of \textit{value}. Besides corner cases, our DSL and templates allow us to effectively implement proper floating point equivalence for problems that return a float. Finally, in many languages, MultiPL-E uses types that are \textit{not} considered best practice, such as in Scala, where it relies on the Java types {\tt ArrayList} instead of the native {\tt List}.

\section{Low-Resource Code Language Models}\label{sec:lang-balance}

Because the data availability can vary greatly by programming language, we can consider the goal of building a multilingual code model as a data-imbalanced multi-task learning problem. Previous work in the multilingual natural language community \cite{Conneau2019UnsupervisedCR,arivazhagan2019massively} and in the program synthesis space \cite{ahmad2021unified} have used sampling strategies relying on temperature-scaling. In this work, we use the Unimax \cite{chung2023unimax} strategy to address this imbalance. The Unimax algorithm assumes that we are given a budget of how many examples we plan to consume during training and a maximum number of times, $N$, any single example can be duplicated in the training corpus. Then, we separate the data into buckets by programming language and add $N$ epochs of each of the lowest-resource languages until we can safely distribute the remaining budget across all the remaining languages without exceeding $N$ epochs over any one of these remaining languages. This will allow us to control the number of epochs $N$ we perform over the low-resource languages to minimize overfitting while allowing fair distribution of the compute budget to the remaining high-resource languages. We will ablate the choice of $N$ in our experiments. 


\begin{figure}[h]
    \centering
    \caption{Different distributions for Unimax with different budgets.}
    \includegraphics[width=0.45\textwidth]{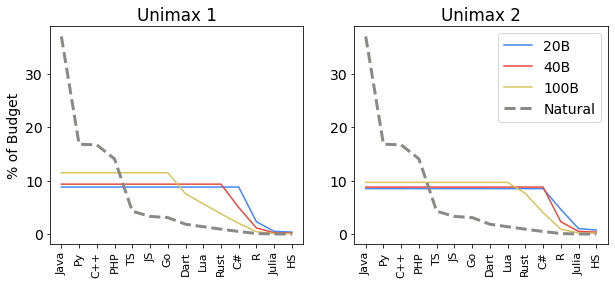}
    \label{fig:distributions}
\end{figure}
\section{Experimental Setup}
\subsection{Models}
To understand the impact of training decoder-only models on the different programming language distributions, we train models in 3 sizes: 1B, 2B, and 4B. For each of these sizes, we train 5 different models on each distribution: Natural and Unimax $N$, where $N\in \{1,2,3,4\}$. The parameters and training differences are listed in \autoref{tab:model-hparams}. We follow \citet{chowdhery2022palm} for all other architecture choices. Every model has a context window of 2048 and is trained identically with the same vocabulary described in \autoref{sec:vocab}. We use a base learning rate of 0.01 and a constant warmup with a step inverse decay. The number of warmup steps is kept to 10\% of the total training steps per model. The total number of training steps is 38000, 77000, 190000 for the 1B, 2B, and 4B models, respectively. We use the Adafactor optimizer  \citep{shazeer2018adafactor} and a batch size of 256. We prepend {\tt [code]} to the beginning and add the tag {\tt [eod]} to the end of each file from our training data. Finally, we use the T5X and SeqIO  \citep{roberts2022t5x} frameworks. We use the UL2 \citep{tay2022unifying} objective with an additional causal language modeling objective as described in \autoref{appendix:train-obj}.

\begin{table}[]
    \centering
    \caption{Hyperparameters for models trained (BC) compared with those used to train PaLM-Coder(PC). For PaLM-Coder, we report the number of code tokens trained on. Each BC model is trained on each of the naturally occurring distributions of the GitHub data and each of the distributions is detailed in \autoref{sec:lang-balance} where $N\in \{1,2,3,4\}$}
    \begin{tabular}{c|c|c|c|c}
    \hline
        &   \# of &  & &Train  \\
        Model  & Layers & Heads & $d_{model}$ & Tokens(B)  \\
    \hline
        BC 1B &  16     & 8          & 8192 & 20.2 \\
        BC 2B &  24     & 16         & 10240 & 40.4   \\       
        BC 4B &26     & 16         & 14336  &  100   \\
        \midrule
        PC 8B & 32 & 16 & 4096 & 46.8   \\
        PC 62B & 64 & 32 & 8192 & 46.8    \\
    \end{tabular}
    \label{tab:model-hparams}
\end{table}
\subsection{Training Data}
Our curated source code corpus was obtained by collecting publicly available code data on the web using a custom code data collection system. We apply a similar license filter as \citet{kocetkov2022stack} to remove any files with non-permissible licenses, use simple heuristics to filter out low-quality code and apply near-deduplication to obtain our corpus of high quality, permissive source code. After preprocessing, we select 14 programming languages by their file extensions according to the mapping used by GitHub's Linguist library\footnote{\href{https://github.com/github/linguist/blob/master/lib/linguist/languages.yml}{https://github.com/github/linguist/}} to segment the dataset by language. To calculate the number of examples per language, we use SeqIO's caching feature and take the number of examples after post-processing  \citep{roberts2022t5x}. We list the percentages of all examples and file extensions used per language in \autoref{appendix:train-langs}. With these numbers, we consider the top 7 languages to be \textbf{high-resource}(HR): Java, Python, C++, PHP, TypeScript, JavaScript, and Go. We further consider the bottom 7 languages to be \textbf{low-resource}(LR): Dart, Lua, Rust, C\#, R, Julia, and Haskell.

\subsection{Vocabulary}\label{sec:vocab}
The original PaLM \cite{chowdhery2022palm} vocabulary focuses on multilingual natural language. In contrast, we trained our SentencePiece \cite{kudo2018sentencepiece} vocabulary with 64k tokens from the training data directly. Each programming language is uniformly sampled to build the vocabulary. In previous works, such as \citet{Chen2021EvaluatingLL}, a list of tokens that consists of a different number of whitespace is manually added to represent code more efficiently. In our work, we rely on the SentencePiece model to learn the whitespace tokens by allowing extra whitespace tokens and whitespace-only tokens. In the end, the model can represent up to 12 whitespaces into one token. In addition, numbers are split into individual tokens.

\subsection{Benchmarks}
BabelCode currently supports 4 datasets. To allow the translation of any dataset to any language, we modify each benchmark as well as remove problems that were incompatible. These changes are described in \autoref{appendix:dataset}. For HumanEval \cite{Chen2021EvaluatingLL}, MBPP \citep{austin2021program}, and Transcoder \citep{roziere2020unsupervised}, we add the prefix \textbf{BabelCode-} (\textbf{BC}) to indicate that we are using the BabelCode specific version. Further, for Transcoder, we use the same version as in \citet{chowdhery2022palm}. \textbf{BC-HumanEval} (\textbf{BC-HE}) has 161 out of the original 164 HumanEval questions. \textbf{BC-MBPP} has 855 of the original 999 questions. \textbf{BC-Transcoder} (\textbf{BC-TC}) has 524 of the original 560 questions.

We additionally introduce a new dataset called \textbf{Translating Python Programming Puzzles (TP3)}. We take the verification functions from the questions in the original Python Programming Puzzles dataset \citep{schuster2021programming} to create this dataset. These functions are hand-crafted by the authors and are used to check if an answer satisfies the constraints of the puzzle. These puzzles range in difficulty from basic character checking to competitive programming problems. Thus, each verification function is written by an expert python programmer and requires a significant understanding of programming to translate. In total, there are 370 python functions to translate. Examples from TP3 can be found in \autoref{appendix:ds-tp3}.

\subsection{Evaluation}

For BC-HumanEval, we follow \citet{Chen2021EvaluatingLL} and generate 200 programs per problem. Further, we use a zero-shot prompt described in \autoref{appendix:gen-prompt}. We use the built-in docstring translation of \babelcode. We generate 50 programs per problem on our three translation tasks and use the prompts described in \autoref{appendix:trans-prompt}. We consider these prompts zero-shot as we do not provide any additional examples. However, we provide the translated signature without the docstring in the prompt. We do not consider this to be data leakage as it is trivial to translate signatures with libraries such as Treesitter\footnote{\href{https://tree-sitter.github.io/tree-sitter/}{https://tree-sitter.githcub.io/tree-sitter/}}. 

For every dataset, we use $T=0.8$, $top_p=0.95$, and do not use $top_k$. We use the $pass@k$ estimator \citep{Chen2021EvaluatingLL} to measure the performance. We use $k=100$ and $k=25$ for generation and translation, respectively.  

\section{Results}\label{sec:results}
\begin{figure*}[h]
\centering
\caption{Comparison of the models trained with PaLM-Coder models. For each dataset, we use \citet{Chen2021EvaluatingLL} $pass@k$ estimator with $n=2*k$. We then generate $n$ samples per problem with $T=0.8$. Full results can be found in \autoref{appendix:results}. Languages in the X-Axis are sorted from high to low resource. HS is Haskell, JS is JavaScript, Py is Python, and TS is TypeScript.}
\includegraphics[width=\textwidth]{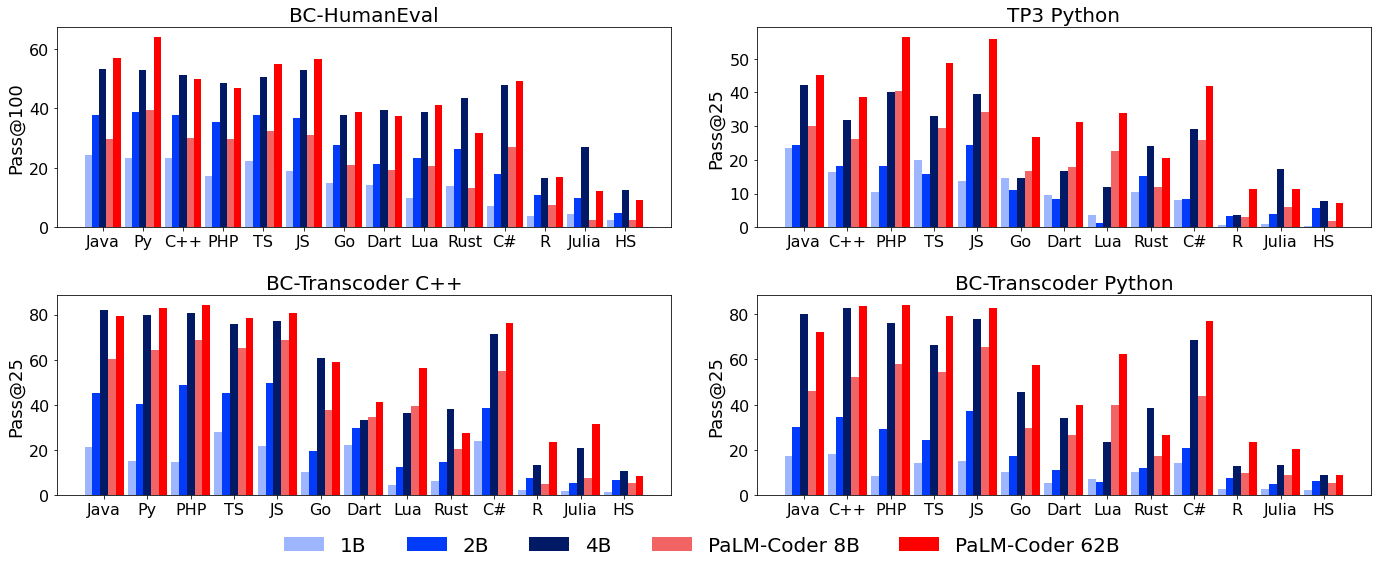}
\label{fig:baselines}
\end{figure*}

\subsection{Baseline Models}
We report the baseline results for our trained models and PaLM-Coder in \autoref{fig:baselines}. On BC-HumanEval, we find that the 2B model has a better $pass@100$ than that of PaLM-Coder 8B on all but C\# and Python. On average, the BC-2B model trained on the natural distribution of GitHub data has average improvements of 48.17\% compared to PaLM-Coder 8B despite having a quarter of the number of parameters and training on 6.4B fewer code tokens. Further, we find that the 4B model outperforms PaLM-Coder 62B on 6 of the 14 languages evaluated. This likely results from the 4B model seeing over 53B tokens more than what PaLM-Coder 62B did. Another likely factor in this discrepancy is that the data PaLM-Coder was fine-tuned on included all languages on GitHub in contrast to our filtered training dataset. 

We also observe that performance on languages do not scale with respect to their resource level nor the model's size. C\#, Dart, Julia, and Haskell have significantly higher gains when scaling to 4B model size when compared to the other languages. While this may be due to the increased number of training tokens, it is not consistent across all LR languages as the increase in performance for R and Lua when scaling from 1B to 2B is similar to that when scaling from 2B to 4B. Instead, this result is likely due to better transfer from languages such as Java, Python, and C++.

The importance of scale for multi-lingual code models is further demonstrated by the results of the translation tasks. We find that in BC-TP3, the 1B and 2B models' performance is similar. However, the most significant gains are from scaling up to 4B where it beats PaLM-Coder 8B on all but three languages in this zero-shot translation. We do make note, though, that while we do not provide any examples for in-context learning, we do provide the signature in the target language during generation. This finding is less pronounced in BC-Transcoder as the scaling observed in all languages is more akin to that seen in BC-HumanEval. 

\subsection{Impact of Balancing Programming Languages}\label{sec:pl-balance}
\begin{figure*}[h]
\centering
\caption{Effects of scale on the average $pass@k$ of the high and low resource languages for each of four datasets. Full tabulated results are located in \autoref{appendix:results}.}
\includegraphics[width=\textwidth]{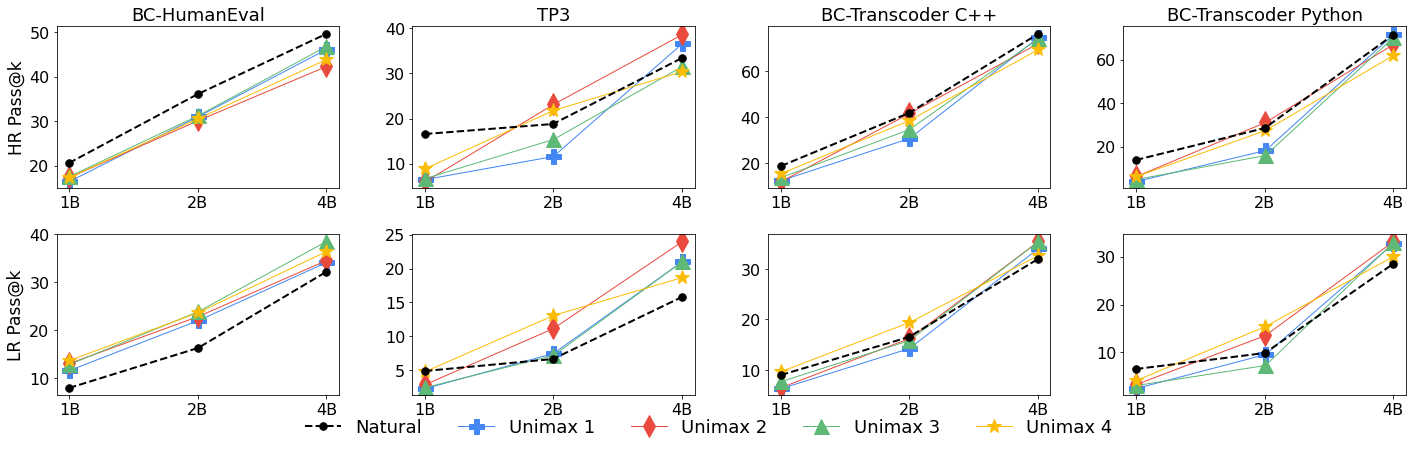}
\label{fig:pk-scaling}
\end{figure*}

\begin{figure*}[h]
\centering
\caption{Mean relative difference of $pass@k$ for each of the models trained on the different Unimax distributions compared to the $pass@k$ of the same sized model trained on the Natural distribution. The X-Axis is the language sorted from high to low resource. HS is Haskell and Py is Python. The percent changes for each delta for HR languages are shown in \autoref{tab:hr-pct-change} and \autoref{tab:lr-pct-change} for LR languages.}
\includegraphics[width=\textwidth]{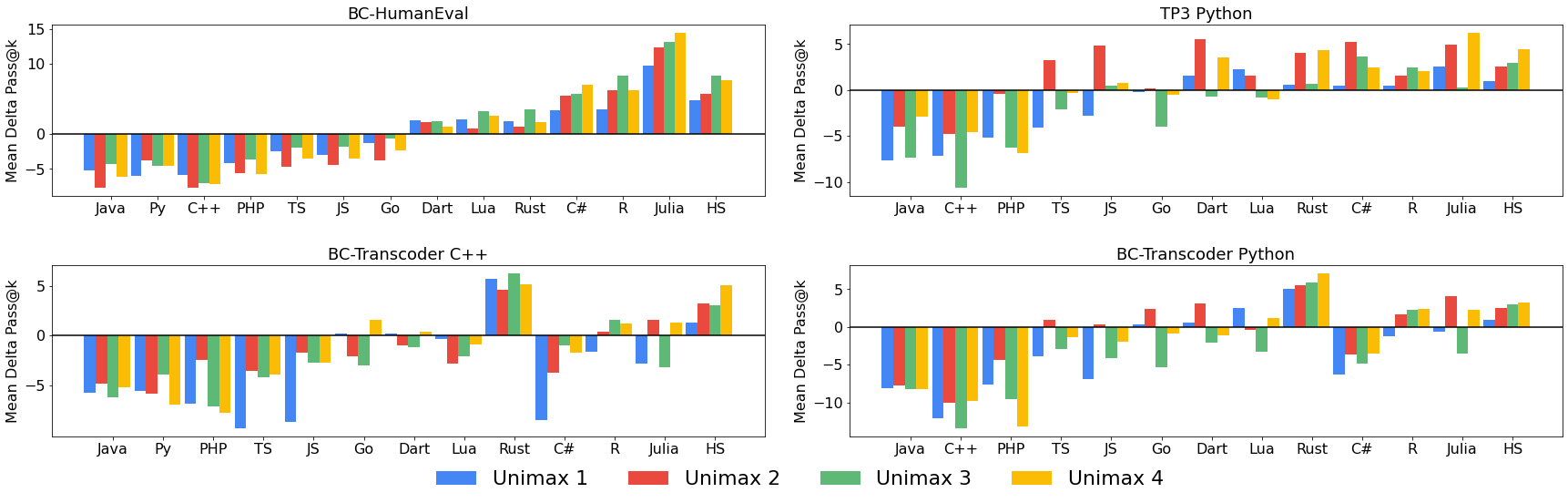}
\label{fig:scale-delta}
\end{figure*}

\autoref{fig:pk-scaling} shows the mean $pass@k$ scores of different models trained on each of the 5 distributions for each of the 4 datasets. As expected, the natural distribution is optimal if the focus is solely HR languages as the performance losses when training on Unimax balanced data are 15.47\%, 14.00\%, and 9.35\% for the 1B, 2B, and 4B models, respectively. However, for any LR language, Unimax is clearly better given that there is an average $pass@100$ improvement on these languages of 111.85\%, 68.38\%, and 19.22\% for the 1B, 2B, and 4B size models, respectively. For generation tasks, we find that $N=3$ is optimal with respect to the difference between performance gained on LR and performance lost on HR languages. On the 1B, 2B, and 4B models, the ones trained on the Unimax 3 dataset had differences of 130.17\%, 87.80\%, and 36.00\%, respectively. 

We observe similar scaling trends on TP3, as training on a Unimax distribution yielded average $pass@25$ improvements to LR languages of 124.45\% for the 1B model, 64.51\% for the 2B model, and 51.29\% for the 4B model when compared to the same sized models trained on the natural distribution. Unlike BC-HumanEval, training the 4B on Unimax Distributions yielded \textit{better} average HR performance with an increase of 6.80\%. As shown in \autoref{fig:scale-delta}, training a 4B model on the Unimax 2 distribution had a mean $pass@25$ improvement of 71.59\% in LR languages and an improvement of 20.31\% on HR languages when compared to the natural distribution. Training on other Unimax distributions does not see as large of improvements. For the 4B model, we find mean LR improvements of 42.39\%, 52.91\%, and 38.26\% when trained on the Unimax 1, 3, and 4 distributions, respectively. This indicates that for TP3, at least, balancing the training data for each language improves translation capabilities. However, less Python data adversely affects understanding the source code necessary to translate it properly.


When evaluated on BC-Transcoder, we find that LR performance \textit{increased} with size. When the source language is C++, training on the Unimax distributions yielded an average $pass@25$ improvements of 7.57\%, 6.76\%, and 11.80\% for the 1B, 2B, and 4B models, respectively. Translating Python to other languages followed this trend with an average change of -26.04\%, 15.1\%, and 22.47\% for the 1B, 2B, and 4B models, respectively. On BC-Transcoder, we find similar benefits when translating from Python to other languages, although the performance on higher resource languages is significantly worse. When translating from C++ to other languages, we find that training both a 1B and 2B model on the UM 4 distribution improves performance on 5 of the 7 LR languages. For 4B sized models, the UM 2 distribution is optimal as LR performance increased by an average of 20.47\% when compared to training on the natural distribution.  As the source code of BC-Transcoder focuses on language-agnostic algorithm implementations, this scaling trend is most likely due to the importance of a surface-level understanding of the target language. Further, the fact that this trend does not appear for BC-HumanEval or TP3 indicates that neither model size nor duplication of language data enables the model to have a deep understanding of these low-resource languages.

\subsection{Effects Of Language Balance on Predictions}
\begin{figure*}[h]
\centering
\caption{Results on BC-HumanEval and BC-TP3 at a prediction level. Left to right: 1) The \% of predictions that passed at least one test, but not all 2) The average, per question, percent of tests passed for each prediction 3) The \% of predictions that had either a compilation error, runtime error, or timed out. Full results for BC-HumanEval and BC-TP3 can be found in \autoref{fig:he-qual} and \autoref{fig:tp3-qual}, respectively.}
\includegraphics[width=\textwidth]{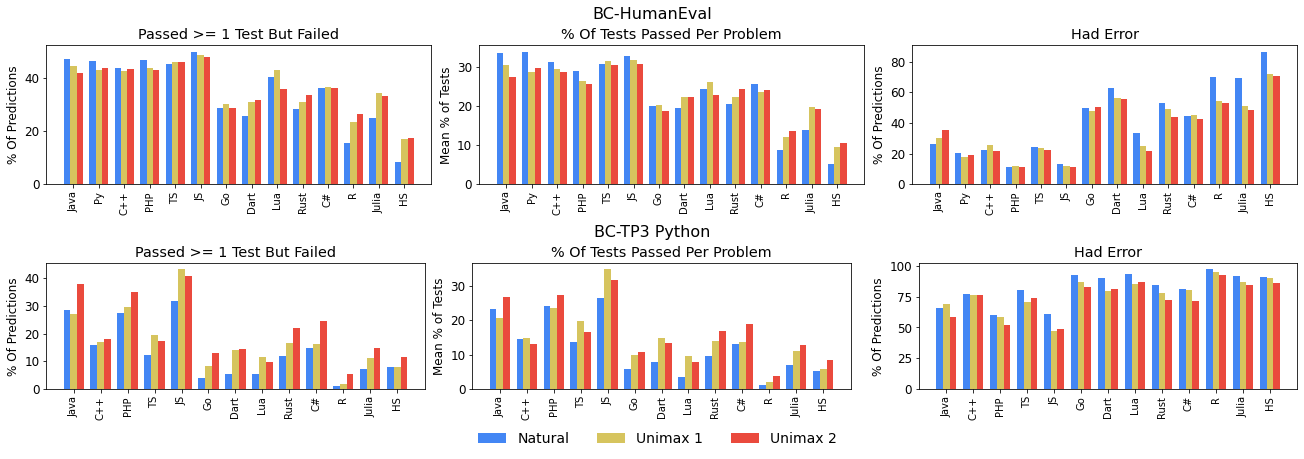}
\label{fig:qual-results}
\end{figure*}

We find that, as is expected, decreasing the number of tokens for a language negatively impacts its performance on that language. To compare the overall effects of language balancing at each size, we focus on the Unimax 1 and Unimax 2 distributions as they represent the largest change in proportions of HR languages when compared to the Natural distribution. \autoref{fig:qual-results} shows that on BC-HumanEval, training on either UM 1 or UM 2 will cause the model to generate fewer correct solutions than when the model is trained on the Natural distribution with respect to HR languages. However, this is \textit{not} due to those models generating more programs with either compilation or run-time errors as the raw average increase is only 0.40 and 1.15 for the models trained on the Unimax 1 and Unimax 2 respectively. Rather, we find that the largest decrease is in the mean \% test cases passed per problem. Training on the Unimax 1 and Unimax 2 distributions results in 5.50\% and 9.09\% fewer test cases respectively when compared to the model trained on the natural distribution.

On LR languages, the Unimax 1 distribution yielded the best improvements compared to the other distributions. Specifically, the programs generated by the model trained on the Natural distribution passed, on average, 5.13\% of the test cases per problem. In comparison, 9.53\% and 10.48\% of average test cases per problem were solved by the models trained on the Unimax 1 and Unimax 2 distributions. The less than 1\% improvement when going from Unimax 1 to Unimax 2 suggests that, for generation tasks, multi-lingual models of code benefit the most from seeing unique data.

In our translation task of TP3, we observe consistent improvements in the mean number of test cases passed for both HR and LR languages. For the former, we observe an average improvement of 2.58\% and 3.06\% compared to the Natural distribution for the UM 1 and 2 distributions respectively. On LR languages, we find average improvements of 3.40\% and 4.99\% over the Natural distribution for the UM 1 and UM 2 distributions respectively. These results, along with the performance improvements discussed in \autoref{sec:pl-balance}, indicate that translation tasks benefit highly from uniformly balanced languages. This is, likely, due to the task formulation where natural language understanding is not necessary. Higher resource languages are more likely to contain diverse natural language and code pairs due to the language's popularity. 

Thus, performance on NL2Code tasks, such as BC-HumanEval, depends on the unique samples of code and doc-strings in the training corpus. Translation, on the other hand, does not have this constraint. Rather, it appears that uniformly balancing languages is the optimal strategy for this task.

\section{Related Works}
\textbf{Code Evaluation } Existing code benchmarks have primarily focused on surface matching evaluation \citep{Lu2021CodeXGLUEAM,Yin2018LearningTM,Wang2022MCoNaLaAB,Husain2019CodeSearchNetCE}. Recent works have introduced new execution-based benchmarks for both generation \citep{austin2021program,hendrycks2021measuring, Chen2021EvaluatingLL,Lai2022DS1000AN} and repair \citep{yasunaga2021break} tasks, however, these have been limited to only Python. Additional works have introduced generation \citep{Li2022CompetitionlevelCG} and translation \citep{roziere2020unsupervised} tasks in multiple-languages, but are limited to only C++, Java, and Python. We acknowledge concurrent works by \citet{cassano2022scalable} and \citet{athiwaratkun2023multilingual} on translating HumanEval and MBPP into multiple programming languages. As we note in \autoref{sec:diff-prior}, \babelcode{} supports deeper analysis on a wider range of tasks while including significant methods for ensuring correctness. 

\textbf{Code LLMs } Recent years has seen significant interest in LLMs for code. CodeBERT \citep{feng-etal-2020-codebert} is the first work to train an encoder only model on code. CodeT5 \citep{wang-etal-2021-codet5}, PLBART \citep{ahmad2021unified}, and additional works \citep{clement-etal-2020-pymt5,Orlanski2021ReadingSE,Chakraborty2022NatGenGP} examine training encoder-decoder models on code. Similar to this work, \citet{ahmad2021unified} investigate difference data balancing strategies for pre-training. Our work differs in that we focus on balancing many programming languages in pre-training data. AlphaCode \citep{Li2022CompetitionlevelCG}, Codex \citep{Chen2021EvaluatingLL}, PaLM \citep{chowdhery2022palm}, and other works \citep{nijkamp2022conversational,Fried2022InCoderAG,allal2023santacoder,Christopoulou2022PanGuCoderPS} have shown that decoder-only code language models achieve exceptional performance on a wide range of tasks. Additional works have investigated different training strategies \citep{roziere2020unsupervised, bavarian2022efficient} and different pre-training data \citep{Rozire2021DOBFAD,Orlanski2022EvaluatingHF,austin2021program}. 

\textbf{Language Balancing} Choosing a proper sampling distribution from a mixture of datasets of various size is a difficult problems. Initial attempts at studying this in the multilingual natural language processing literature relied on temperature-based approaches \cite{Conneau2019UnsupervisedCR,arivazhagan2019massively}. These approaches oversample the low-resource tasks and downsample the high-resource ones. Other works have adopted more dynamic approaches, adapting the sampling rates in an online fashion during training \cite{wang2020balancing}. 

\section{Conclusion}

We proposed the \babelcode{} framework for multi-lingual execution-based evaluation and a new strategy for balancing programming language distributions. We highlight the ease of creating new benchmarks with \babelcode{} by proposing the Translating Python Programming Puzzles. Our experiments demonstrate that adjusting how much we oversample low-resource languages and downsample high-resource languages greatly improves low-resource performance with minimal impact to to the performance of high-resource languages in tasks involving either a single or multiple programming language. By open-sourcing \babelcode{}, future work can investigate improved balancing strategies along with new multi-lingual programming language questions.

\section*{Acknowledgements}
We thank Michael Janner, Owen Lewis, Alex Polozov, Uros Popovic, Devjeet Roy, Tal Schuster, and Charles Sutton for their helpful discussions and feedback on the paper. 
\bibliography{references}
\bibliographystyle{icml2023}

\newpage
\appendix
\onecolumn
\section{BabelCode Design}\label{appendix:bc-design}
\begin{figure*}[h]
\centering
\caption{Sample problem translated from Python to C++ using \babelcode{}}
\includegraphics[width=1\textwidth]{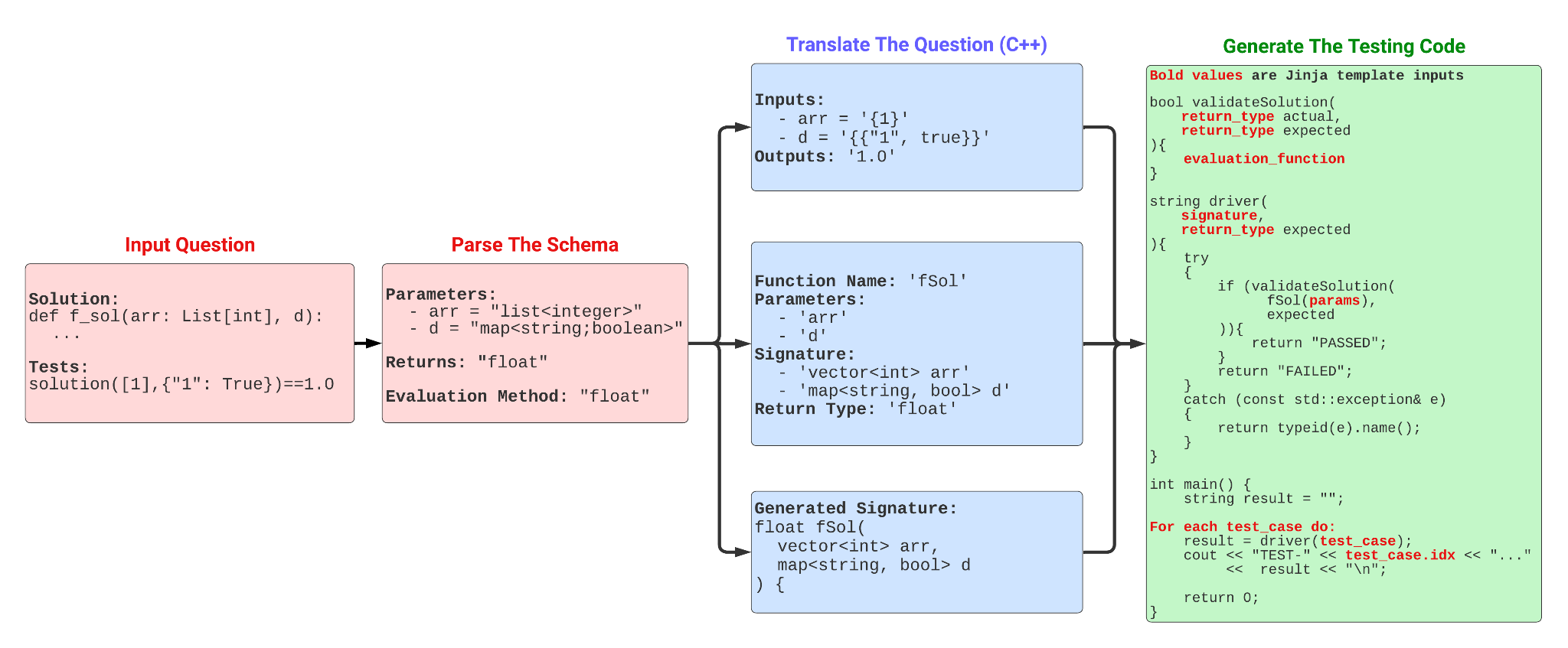}
\label{fig:bc-sample}
\end{figure*}
\babelcode{}'s design shares many similarities to \citet{athiwaratkun2023multilingual} and \citet{cassano2022scalable}. For translation, we too implement a recursive visitor pattern to translate input and output values to the corresponding code in the target language. When converting a coding dataset, we follow prior works by parsing {\tt assert} statements using AST parsing libraries to determine the inputs and outputs for a given question. To find the function name for a problem, we once again use AST parsers to find the function definition located in the ground truth solution. The found tree is additionally used for parsing the argument names and types. If the types for either the arguments or returns do not exist, we infer them based on the types found from the literal values of the inputs and outputs. While our implementation differs, the overall process is similar to \citet{athiwaratkun2023multilingual} and \citet{cassano2022scalable}. Following \citet{cassano2022scalable}, we execute the generated code through the command line using each language's recommended commands to compile and run a given script. As \citet{athiwaratkun2023multilingual} is not open sourced, we cannot compare the similarities of this portion.
\section{Dataset Changes}\label{appendix:dataset}
\subsection{Incompatible Problems}\label{appendix:incompatible}

\begin{lstlisting}[language=Python]
def encode_cyclic(s: str):
    """
    returns encoded string by cycling groups of three characters.
    """
    # split string to groups. Each of length 3.
    groups = [s[(3 * i):min((3 * i + 3), len(s))] for i in range((len(s) + 2) // 3)]
    # cycle elements in each group. Unless group has fewer elements than 3.
    groups = [(group[1:] + group[0]) if len(group) == 3 else group for group in groups]
    return "".join(groups)


def decode_cyclic(s: str):
    return encode_cyclic(encode_cyclic(s))

from random import randint, choice
import string
letters = string.ascii_lowercase
for _ in range(100):
    str = ''.join(choice(letters) for i in range(randint(10, 20)))
    encoded_str = encode_cyclic(str)
    assert decode_cyclic(encoded_str) == str
\end{lstlisting}

\subsection{Changes To HumanEval}\label{appendix:ds-humaneval}
Original:

\begin{lstlisting}[language=Python]
def reverse_delete(s,c):
    """ Task
    We are given two strings s and c, you have to deleted all the characters in s that are equal to any character in c
    then check if the result string is palindrome.
    A string is called palindrome if it reads the same backward as forward.
    You should return a tuple containing the result string and True/False for the check.
    Example
    For s = "abcde", c = "ae", the result should be ('bcd',False)
    For s = "abcdef", c = "b"  the result should be ('acdef',False)
    For s = "abcdedcba", c = "ab", the result should be ('cdedc',True)
    """
    s = ''.join([char for char in s if char not in c])
    return (s,s[::-1] == s)

assert reverse_delete('abcde', 'ae') == ('bcd', False)
assert reverse_delete('abcdef', 'b') == ('acdef', False)
assert reverse_delete('abcdedcba', 'ab') == ('cdedc', True)
\end{lstlisting}

Modified:

\begin{lstlisting}[language=Python]
def reverse_delete(s,c):
    """ Task
    We are given two strings s and c, you have to deleted all the characters in s that are equal to any character in c
    then check if the result string is palindrome.
    A string is called palindrome if it reads the same backward as forward.
    You should return a two element list containing the result string and "True" if the check passed, otherwise "False".
    Example
    For s = "abcde", c = "ae", the result should be ('bcd',False)
    For s = "abcdef", c = "b"  the result should be ('acdef',False)
    For s = "abcdedcba", c = "ab", the result should be ('cdedc',True)
    """
    s = ''.join([char for char in s if char not in c])
    return [s,str(s[::-1] == s)]

assert reverse_delete('abcde', 'ae') == ['bcd', 'False']
assert reverse_delete('abcdef', 'b') == ['acdef', 'False']
assert reverse_delete('abcdedcba', 'ab') == ['cdedc', 'True']
\end{lstlisting}

\subsection{Changes To Transcoder}\label{appendix:ds-transcoder}
Original:
\begin{lstlisting}[language=C++]
int difference_between_highest_and_least_frequencies_in_an_array ( int arr [ ], int n ) {
  sort ( arr, arr + n );
  int count = 0, max_count = 0, min_count = n;
  for ( int i = 0;
  i < ( n - 1 );
  i ++ ) {
    if ( arr [ i ] == arr [ i + 1 ] ) {
      count += 1;
      continue;
    }
    else {
      max_count = max ( max_count, count );
      min_count = min ( min_count, count );
      count = 0;
    }
  }
  return ( max_count - min_count );
}
\end{lstlisting}
Modified:
\begin{lstlisting}[language=C++]
int difference_between_highest_and_least_frequencies_in_an_array(vector<int> arr, int n) {
  sort(arr.begin(), arr.end());
  int count = 0, max_count = 0, min_count = n;
  for ( int i = 0;
  i < ( n - 1 );
  i ++ ) {
    if ( arr [ i ] == arr [ i + 1 ] ) {
      count += 1;
      continue;
    }
    else {
      max_count = max ( max_count, count );
      min_count = min ( min_count, count );
      count = 0;
    }
  }
  return ( max_count - min_count );
}

\end{lstlisting}
\subsection{TP3 Examples}\label{appendix:ds-tp3}
\begin{lstlisting}[language=Python]
def sat(inds: List[int], string):
    return inds == sorted(inds) and ''.join((string[i] for i in inds)) == 'intelligent'
    
assert sat([-10, -5, -1, 0, 2, 2, 3, 4, 7, 8, 12], 'enlightenment') == True
assert sat([-11, -10, -8, -6, -4, -4, -3, -2, -1, 1, 3], 'inntGetlige') == True
assert sat([-10, -5, -1, 0, 2, 2, 3, 4, 7, 8, 12], '  einliJSgeteq ne CAlti') == False

\end{lstlisting}
\section{Training Languages}\label{appendix:train-langs}
\begin{table}[h]
    \centering
    \caption{Languages used for training and the extensions we used to filter files. The percentages of the data are calculated after caching and postprocessing using SeqIO.}
    \begin{tabular}{l|c|c}
    \toprule
       Language &                                         Extensions & \% Of Data \\
    \midrule
             C\# &                    .cs, .cake, .csx, .linq &     0.49\% \\
            \multirow{2}{*}{C++} &  .cpp, .c++, .cc, .cp, .cxx, .h, .h++, .hh, &    \multirow{2}{*}{16.68\%} \\
                & .hpp, .hxx, .inl, .ino, .ipp, .ixx, .re, .tcc, .tpp & \\
           Dart &                                            .dart &     1.85\% \\
             Go &                                              .go &     3.09\% \\
        Haskell &                          .hs, .hs-boot, .hsc &     0.02\% \\
           Java &                            .java, .jav, .jsh &    36.95\% \\
     JavaScript &                              .js, .cjs, .mjs &     3.31\% \\
          Julia &                                              .jl &     0.03\% \\
            Lua &                                             .lua &     1.39\% \\
            \multirow{2}{*}{PHP} &  .php, .aw, .ctp, .fcgi, .inc, .php3,  &    \multirow{2}{*}{14.05\%} \\
            & .php4, .php5, .phps, .phpt & \\
         Python &              .py, .py3, .pyi, .pyw, .pxi &    16.80\% \\
              R &                                .r, .rd, .rsx &     0.11\% \\
           Rust &                                    .rs, .rs.in &     0.93\% \\
     TypeScript &                              .ts, .cts, .mts &     4.28\% \\
    \bottomrule
    \end{tabular}
    \label{tab:lang-exts}
\end{table}
\section{Training Objective}\label{appendix:train-obj}
This paper uses a variant of the UL2 objective \cite{tay2022unifying} for training the code language models. The UL2 objective consists of a mixture of span corruption and prefix language modeling objectives, as defined in \citet{raffel2020exploring}. In this work, we select two span corruption instances using the implementation provided in the T5 library. \footnote{See https://github.com/google-research/text-to-text-transfer-transformer/blob/main/t5/data/preprocessors.py\#L1923} The only differences between these two instances consist of different values for the \texttt{noise\_density} and \texttt{mean\_noise\_span\_length} arguments. In particular,  we use (3.0, 0.15) and (32, 0.5) for  the (\texttt{noise\_density}, \texttt{mean\_noise\_span\_length}) arguments for each span corruption instance respectively. 

The prefix language modeling objective randomly breaks text into two pieces, and the model is tasked to reconstruct the latter, given the former. Finally, we add an additional objective which consists of causal language modeling, which can be considered a special case of prefix language modeling; the first piece consists of the empty string. We assign the probabilities 10\%, 10\%, 20\%, and 60\% for each objective, respectively. 
\section{Prompts Used}\label{appendix:prompts-used}
\subsection{Generation Tasks}\label{appendix:gen-prompt}
\begin{lstlisting}
You are an expert |{\color{red}\{\{ Language \}\}}| programmer, complete the implementation.
Solution in |{\color{red}\{\{ Language \}\}}|:
[BEGIN]

|{\color{red}\{\{ Signature With Docstring \}\}|
\end{lstlisting}
Each {\tt \{\{$\ldots$\}\}} represents a field that is filled in.

Example from HumanEval for generating C\# code:

\begin{lstlisting}
You are an expert C# programmer, complete the implementation.
Solution in C#:
[BEGIN]

class Solution {
    /**
     * Return length of given string
     * >>> GetStringLength("")
     * 0
     * >>> GetStringLength("abc")
     * 3
     */
    public int GetStringLength(string s) {
\end{lstlisting}

\subsection{Translation Tasks}\label{appendix:trans-prompt}
\begin{lstlisting}
Translate the following |{\color{blue}\{\{ Source Language \}\}}| program to |{\color{red}\{\{ Target Language \}\}}|:
Input:

|{\color{blue}\{\{ Source Code \}\}}|

|{\color{red}\{\{ Target Language \}\}}| Translation:
[BEGIN]

|{\color{red}\{\{ Target Signature \}\}}|

\end{lstlisting}

Each {\tt \{\{$\ldots$\}\}} represents a field that is filled in. The {\tt \color{blue} \{\{fields\}\}} correspond to the source language we are translating from, while {\tt \color{red} \{\{fields\}\}} correspond to the target language to translate too.

Example For TP3 translation from Python to Haskell:
\begin{lstlisting}
Translate the following Python program to Haskell:
Input:

def sat(i: int) -> bool:
    return i % 123 == 4 and i > 10 ** 10

Haskell Translation:
[BEGIN]

sat :: Integer -> Bool
sat i = 
\end{lstlisting}

\begin{figure*}[h]
\centering
\caption{Qualitative Comparison of the 4B model trained on the Natural, the Unimax 1, and Unimax 2 distributions when evaluated on BC-HumanEval. The results can be found in \autoref{appendix:he-qual-hr} and \autoref{appendix:he-qual-lr}.}
\includegraphics[width=1\textwidth]{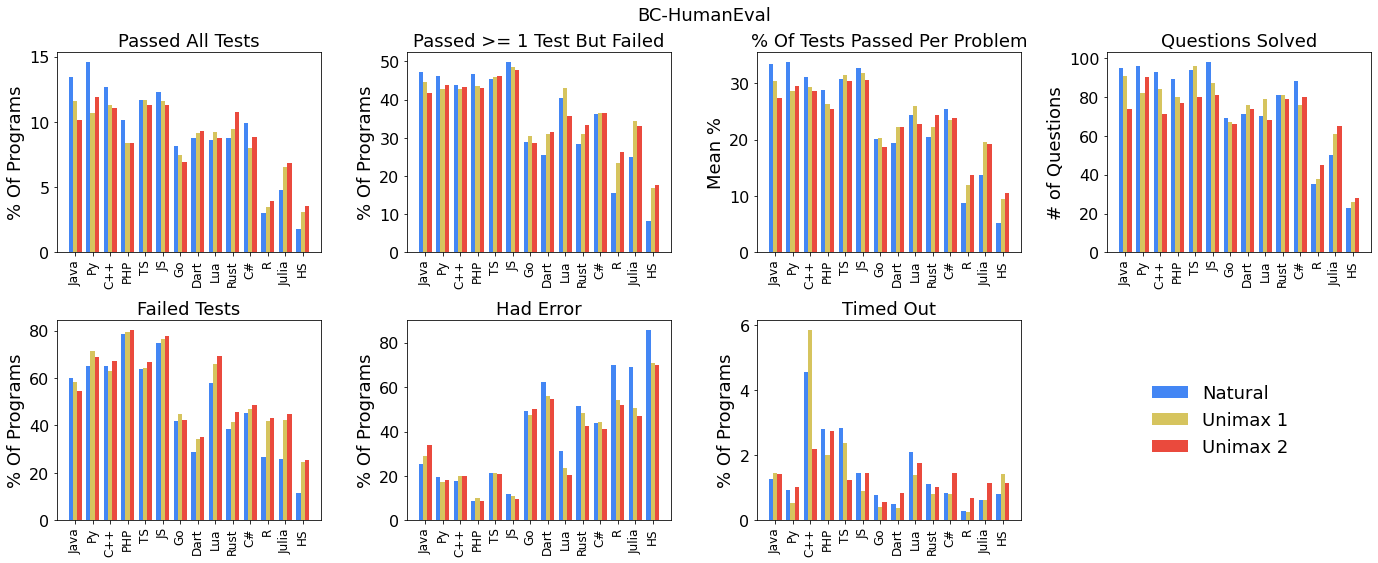}
\label{fig:he-qual}
\end{figure*}

\begin{figure*}[h]
\centering
\caption{Qualitative Comparison of the 4B model trained on the Natural, the Unimax 1, and Unimax 2 distributions when evaluated on TP3. The results can be found in \autoref{appendix:tp3-qual-hr} and \autoref{appendix:tp3-qual-lr}.}
\includegraphics[width=1\textwidth]{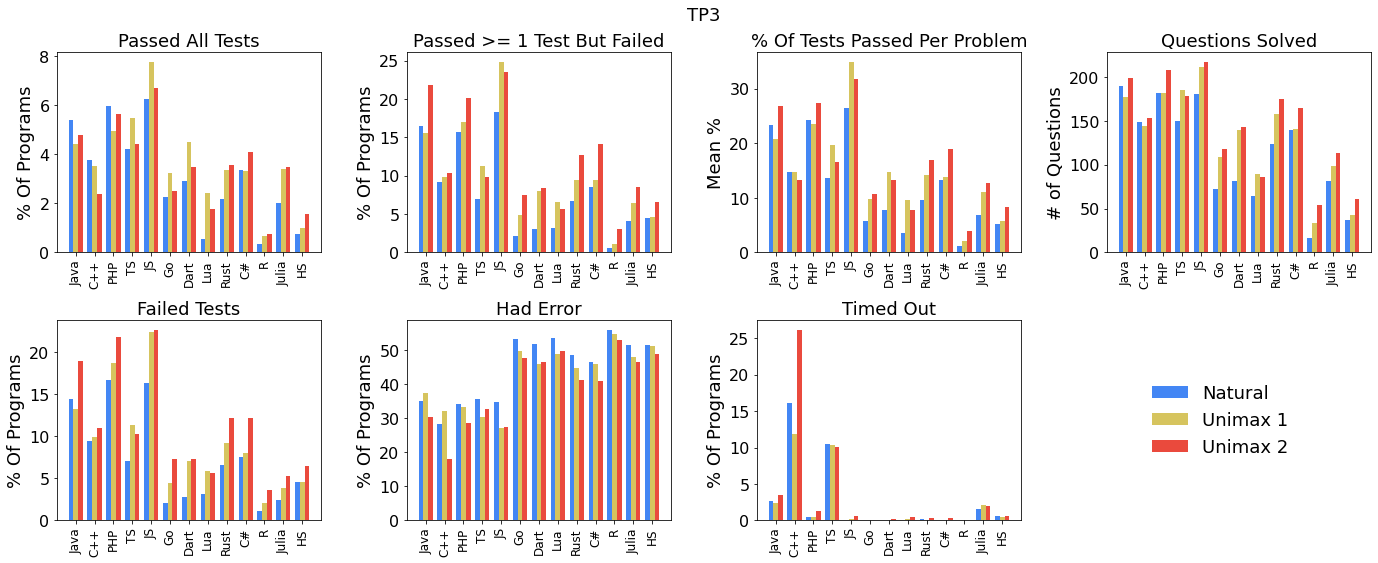}
\label{fig:tp3-qual}
\end{figure*}
\newpage
\section{Full Results}\label{appendix:results}
\begin{table*}[h]
\centering
\caption{BC-HumanEval $pass@1$ values for the different models and training distributions. Used $T=0.8$ and sampled 200 programs per problem. UM is Unimax distribution. PaLM-C is the PaLM-Coder distribution. HS is Haskell, JS is JavaSript, Py is Python, and TS is TypeScript.}
\begin{tabular}{ll|cccccccccccccc}
\toprule
Size & Dist. &    C\# &   C++ &  Dart &   Go &   HS &  Java &    JS &  Julia &  Lua &   PHP &    Py &    R &  Rust &    TS \\
\midrule
\multirow{5}{*}{1B} & Nat &   1.0 &   3.6 &   2.3 &  2.5 &  0.7 &   3.8 &   3.6 &    0.5 &  1.8 &   2.8 &   4.8 &  0.5 &   1.6 &   4.0 \\
    & UM 1 &   1.7 &   3.0 &   3.0 &  2.6 &  1.3 &   2.8 &   4.0 &    2.1 &  2.2 &   2.5 &   3.9 &  1.2 &   2.8 &   4.4 \\
    & UM 2 &   2.0 &   3.2 &   3.0 &  2.7 &  1.6 &   2.7 &   3.9 &    2.1 &  2.1 &   2.3 &   4.2 &  1.4 &   3.1 &   4.3 \\
    & UM 3 &   1.6 &   1.5 &   2.6 &  2.6 &  1.4 &   2.8 &   4.0 &    2.5 &  2.2 &   2.2 &   4.0 &  1.8 &   2.6 &   4.1 \\
    & UM 4 &   1.7 &   2.7 &   3.1 &  2.9 &  1.5 &   2.8 &   3.7 &    2.6 &  2.2 &   2.2 &   3.5 &  2.1 &   2.5 &   4.1 \\
\cline{1-16}
\multirow{5}{*}{2B} & Nat &   2.6 &   7.5 &   5.0 &  5.4 &  1.0 &   8.0 &   7.6 &    1.2 &  4.5 &   6.2 &   9.1 &  1.4 &   3.9 &   7.9 \\
    & UM 1 &   5.3 &   6.0 &   6.1 &  5.1 &  1.9 &   6.6 &   7.6 &    4.4 &  5.4 &   5.6 &   7.8 &  2.1 &   6.4 &   7.5 \\
\bottomrule
\end{tabular}
\label{tab:bc-he-p1}
\end{table*}

\begin{table*}[h]
\centering
\caption{BC-TP3 $pass@1$ values for the different models and training distributions. Used $T=0.8$ and sampled 50 programs per problem. Nat is the natural distribution. UM is Unimax distribution. PaLM-C is the PaLM-Coder distribution. HS is Haskell, JS is JavaSript, Py is Python, and TS is TypeScript.}
\begin{tabular}{ll|cccccccccccccc}
\toprule
Size & Dist.&   C\# &  C++ &  Dart &   Go &   HS &  Java &    JS &  Julia &  Lua &   PHP &    R &  Rust &    TS \\
\midrule
\multirow{5}{*}{1B} & Nat &  0.5 &  1.1 &   0.6 &  1.0 &  0.0 &   1.7 &   1.2 &    0.1 &  0.2 &   0.6 &  0.0 &   0.6 &   2.1 \\
    & UM 1 &  0.1 &  0.2 &   0.1 &  0.5 &  0.2 &   0.3 &   0.7 &    0.1 &  0.1 &   0.1 &  0.0 &   0.3 &   0.7 \\
    & UM 2 &  0.3 &  0.1 &   0.1 &  0.3 &  0.1 &   0.4 &   1.3 &    0.3 &  0.0 &   0.1 &  0.0 &   0.4 &   1.0 \\
    & UM 3 &  0.2 &  0.1 &   0.1 &  0.3 &  0.2 &   0.7 &   1.1 &    0.1 &  0.0 &   0.0 &  0.0 &   0.3 &   0.7 \\
    & UM 4 &  0.2 &  0.3 &   0.4 &  0.3 &  0.8 &   0.6 &   1.1 &    0.7 &  0.1 &   0.2 &  0.0 &   0.7 &   2.1 \\
\cline{1-15}
\multirow{5}{*}{2B} & Nat &  1.0 &  2.2 &   1.3 &  1.9 &  0.8 &   2.9 &   4.1 &    0.3 &  0.1 &   2.8 &  0.4 &   2.2 &   3.1 \\
    & UM 1 &  1.3 &  0.7 &   0.7 &  0.7 &  0.5 &   1.9 &   1.0 &    0.3 &  0.3 &   1.2 &  0.1 &   1.1 &   0.4 \\
\bottomrule
\end{tabular}
\label{tab:bc-tp3-p1}
\end{table*}

\begin{table*}[h]
\centering
\caption{BC-Transcoder with Python source $pass@1$ values for the different models and training distributions where the source language is Python. Used $T=0.8$ and sampled 50 programs per problem. Nat is the natural distribution. UM is Unimax distribution. PaLM-C is the PaLM-Coder distribution. HS is Haskell, JS is JavaSript, Py is Python, and TS is TypeScript.}
\begin{tabular}{ll|cccccccccccccc}
\toprule
Size & Dist. &    C\# &   C++ &  Dart &    Go &   HS &  Java &    JS &  Julia &   Lua &   PHP &    R &  Rust &    TS \\
\midrule
\multirow{5}{*}{1B} & Nat &   1.7 &   2.1 &   0.4 &   1.3 &  0.2 &   2.2 &   2.0 &    0.1 &   0.6 &   0.8 &  0.2 &   1.0 &   2.0 \\
    & UM 1 &   0.1 &   0.1 &   0.0 &   0.4 &  0.2 &   0.3 &   0.5 &    0.0 &   0.0 &   0.1 &  0.1 &   0.8 &   0.8 \\
    & UM 2 &   0.3 &   0.1 &   0.1 &   0.3 &  0.4 &   0.6 &   1.3 &    0.2 &   0.0 &   0.1 &  0.2 &   0.7 &   1.1 \\
    & UM 3 &   0.4 &   0.3 &   0.0 &   0.1 &  0.3 &   0.5 &   0.9 &    0.1 &   0.0 &   0.1 &  0.2 &   0.7 &   0.7 \\
    & UM 4 &   0.3 &   0.2 &   0.2 &   0.1 &  1.2 &   0.6 &   1.1 &    0.2 &   0.1 &   0.4 &  0.3 &   0.9 &   1.3 \\
\cline{1-15}
\multirow{5}{*}{2B} & Nat &   2.9 &   5.5 &   1.0 &   4.4 &  1.0 &   4.9 &   8.2 &    0.3 &   0.4 &   3.8 &  1.3 &   3.5 &   5.2 \\
    & UM 1 &   2.9 &   2.6 &   0.8 &   1.2 &  0.9 &   3.8 &   2.5 &    0.1 &   0.4 &   1.5 &  0.8 &   1.8 &   1.0 \\
\bottomrule
\end{tabular}
\label{tab:bc-tcpy-p1}
\end{table*}

\begin{table*}[h]
\centering
\caption{BC-Transcoder with C++ Source $pass@1$ values for the different models and training distributions. Used $T=0.8$ and sampled 50 programs per problem. Nat is the natural distribution. UM is Unimax distribution. PaLM-C is the PaLM-Coder distribution. HS is Haskell, JS is JavaSript, Py is Python, and TS is TypeScript.}
\begin{tabular}{ll|cccccccccccccc}
\toprule
Size & Dist. &    C\# &  Dart &    Go &   HS &  Java &    JS &  Julia &   Lua &   PHP &    Py &    R &  Rust &    TS \\
\midrule
\multirow{5}{*}{1B} & Nat &   3.0 &   3.1 &   1.3 &  0.1 &   2.6 &   2.3 &    0.1 &   0.3 &   1.0 &   1.8 &  0.1 &   0.6 &   3.7 \\
    & UM 1 &   0.4 &   3.1 &   0.9 &  0.2 &   1.1 &   1.1 &    0.0 &   0.1 &   0.8 &   2.3 &  0.2 &   0.8 &   1.9 \\
    & UM 2 &   1.1 &   1.6 &   0.5 &  0.4 &   1.1 &   2.7 &    0.0 &   0.0 &   1.1 &   1.5 &  0.1 &   0.7 &   1.9 \\
    & UM 3 &   1.9 &   1.6 &   0.8 &  0.4 &   1.6 &   2.9 &    0.0 &   0.0 &   0.7 &   2.3 &  0.1 &   1.1 &   1.5 \\
    & UM 4 &   1.3 &   3.7 &   1.8 &  1.3 &   1.7 &   3.0 &    0.2 &   0.5 &   2.4 &   1.8 &  0.3 &   1.6 &   3.7 \\
\cline{1-15}
\multirow{5}{*}{2B} & Nat &   8.9 &  13.3 &   5.5 &  1.2 &   9.2 &  16.0 &    0.3 &   1.5 &  12.4 &  11.2 &  1.4 &   4.6 &  12.1 \\
    & UM 1 &   4.1 &   7.9 &   3.4 &  1.4 &   6.7 &   6.2 &    0.2 &   2.1 &   5.0 &   6.8 &  0.5 &   3.6 &   4.3 \\
\bottomrule
\end{tabular}
\label{tab:bc-tccpp-p1}
\end{table*}

\begin{table*}[h]
\centering
\caption{BC-HumanEval $pass@100$ values for the different models and training distributions. Used $T=0.8$ and sampled 200 programs per problem. Nat is the natural distribution. UM is Unimax distribution. PaLM-C is the PaLM-Coder distribution. HS is Haskell, JS is JavaSript, Py is Python, and TS is TypeScript.}
\begin{tabular}{ll|cccccccccccccc}
\toprule
Size & Dist. &  C\# &   C++ &  Dart &    Go &    HS &  Java &    JS &  Julia &   Lua &   PHP &    Py &     R &  Rust &    TS \\
\midrule
\multirow{5}{*}{1B} & Nat &   7.3 &  23.2 &  14.4 &  14.9 &   2.4 &  24.3 &  19.0 &    4.4 &   9.8 &  17.1 &  23.3 &   4.0 &  13.9 &  22.4 \\
   & UM 1 &  12.3 &  16.2 &  14.0 &  12.0 &   7.5 &  17.3 &  18.2 &   13.0 &  13.1 &  15.0 &  19.9 &   7.9 &  14.5 &  17.8 \\
   & UM 2 &  14.5 &  16.9 &  13.8 &  11.9 &   8.3 &  19.6 &  19.1 &   15.8 &  13.5 &  14.8 &  21.2 &  10.4 &  16.5 &  19.1 \\
   & UM 3 &  13.7 &  13.5 &  13.4 &  15.4 &  10.0 &  21.4 &  18.4 &   14.2 &  12.8 &  14.6 &  21.1 &  10.4 &  16.0 &  18.6 \\
   & UM 4 &  15.8 &  16.6 &  13.8 &  12.3 &   9.7 &  19.7 &  18.1 &   16.6 &  14.3 &  15.3 &  20.6 &  10.6 &  15.9 &  19.6 \\
\cline{1-16}
\multirow{5}{*}{2B} & Nat &  17.9 &  37.8 &  21.3 &  27.8 &   4.9 &  37.8 &  36.8 &    9.7 &  23.3 &  35.3 &  38.8 &  10.9 &  26.5 &  37.9 \\
   & UM 1 &  28.5 &  31.8 &  24.6 &  26.2 &  12.2 &  32.0 &  33.8 &   23.8 &  22.9 &  29.3 &  30.9 &  14.0 &  29.9 &  34.9 \\
\bottomrule
\end{tabular}
\label{tab:bc-he-p100}
\end{table*}

\begin{table*}[h]
\centering
\caption{BC-TP3 $pass@25$ values for the different models and training distributions where the source language is Python. Used $T=0.8$ and sampled 50 programs per problem. Nat is the natural distribution. UM is Unimax distribution. PaLM-C is the PaLM-Coder distribution. HS is Haskell, JS is JavaSript, Py is Python, and TS is TypeScript.}
\begin{tabular}{ll|cccccccccccccc}
\toprule
Size & Dist.       &    C\# &   C++ &  Dart &    Go &    HS &  Java &    JS &  Julia &   Lua &   PHP &     R &  Rust &    TS \\
\midrule
\multirow{5}{*}{1B} & Nat &   8.2 &  16.5 &   9.6 &  14.6 &   0.3 &  23.4 &  13.8 &    1.1 &   3.7 &  10.6 &   0.8 &  10.6 &  19.8 \\
   & UM 1 &   2.3 &   3.2 &   2.8 &   9.0 &   2.6 &   6.5 &   9.3 &    2.0 &   1.1 &   1.6 &   0.2 &   5.5 &   9.8 \\
   & UM 2 &   5.2 &   1.3 &   1.7 &   5.4 &   1.7 &   8.2 &   9.8 &    4.3 &   1.2 &   1.1 &   0.4 &   6.2 &  11.3 \\
   & UM 3 &   4.5 &   2.1 &   2.3 &   4.4 &   3.7 &  12.6 &  11.6 &    1.2 &   0.4 &   0.4 &   0.1 &   5.4 &   9.2 \\
   & UM 4 &   3.4 &   4.6 &   5.9 &   5.2 &   5.4 &  10.8 &  11.4 &    8.6 &   1.7 &   4.6 &   0.7 &   8.3 &  17.0 \\
\cline{1-15}
\multirow{5}{*}{2B} & Nat &   8.3 &  18.2 &   8.3 &  11.1 &   5.7 &  24.5 &  24.4 &    3.8 &   1.4 &  18.3 &   3.4 &  15.3 &  15.7 \\
   & UM 1 &  15.8 &  11.9 &   6.3 &   8.7 &   5.2 &  23.2 &  10.9 &    5.4 &   4.6 &  11.1 &   2.1 &  13.7 &   4.6 \\
\bottomrule
\end{tabular}
\label{tab:bc-tp3-p25}
\end{table*}

\begin{table*}[h]
\centering
\caption{BC-Transcoder $pass@25$ values for the different models and training distributions where the source language is Python. Used $T=0.8$ and sampled 50 programs per problem. Nat is the natural distribution. UM is Unimax distribution. PaLM-C is the PaLM-Coder distribution. HS is Haskell, JS is JavaSript, Py is Python, and TS is TypeScript.}
\begin{tabular}{ll|cccccccccccccc}
\toprule
    Size & Dist.        &    C\# &   C++ &  Dart &    Go &    HS &  Java &    JS &  Julia &   Lua &   PHP &     R &  Rust &    TS \\
\midrule
\multirow{5}{*}{1B} & Nat &  14.0 &  18.4 &   5.4 &  10.3 &   2.1 &  17.3 &  15.3 &    2.8 &   7.4 &   8.7 &   2.9 &  10.3 &  14.4 \\
    & UM 1 &   3.1 &   1.3 &   1.1 &   5.7 &   3.3 &   4.9 &   5.5 &    0.9 &   1.0 &   1.3 &   1.8 &   6.7 &   7.6 \\
    & UM 2 &   4.3 &   2.7 &   2.2 &   3.5 &   3.9 &   8.2 &  11.0 &    3.3 &   0.4 &   2.9 &   2.6 &   6.9 &  10.4 \\
    & UM 3 &   5.8 &   5.1 &   0.8 &   2.1 &   3.9 &   6.8 &   9.4 &    1.7 &   0.9 &   1.2 &   2.6 &   6.3 &   7.2 \\
    & UM 4 &   5.1 &   3.8 &   2.6 &   1.0 &   6.6 &   7.6 &  11.4 &    2.6 &   1.3 &   5.4 &   3.6 &   7.8 &  10.6 \\
\cline{1-15}
\multirow{5}{*}{2B} & Nat &  20.9 &  34.7 &  11.0 &  17.5 &   6.3 &  30.0 &  37.0 &    5.0 &   5.9 &  29.4 &   7.6 &  11.9 &  24.3 \\
    & UM 1 &  21.4 &  22.3 &  10.8 &  12.5 &   5.2 &  27.7 &  23.0 &    2.5 &   5.6 &  15.9 &   6.4 &  15.2 &  10.9 \\
\bottomrule
\end{tabular}
\label{tab:bc-tcpy-p25}
\end{table*}

\begin{table*}[h]
\centering
\caption{BC-Transcoder $pass@25$ values for the different models and training distributions where the source language is C++. Used $T=0.8$ and sampled 50 programs per problem. Nat is the natural distribution. UM is Unimax distribution. PaLM-C is the PaLM-Coder distribution. HS is Haskell, JS is JavaSript, Py is Python, and TS is TypeScript.}
\begin{tabular}{ll|cccccccccccccc}
\toprule
Size & Dist.    &        &    C\# &  Dart &    Go &    HS &  Java &    JS &  Julia &   Lua &   PHP &    Py &     R &  Rust &    TS \\
\midrule
\multirow{5}{*}{1B} & Nat &  24.2 &  22.4 &  10.4 &   1.5 &  21.2 &  22.0 &    1.9 &   4.7 &  14.9 &  15.4 &   2.3 &   6.4 &  27.9 \\
    & UM 1 &   8.1 &  21.4 &   8.1 &   2.6 &  14.4 &  12.3 &    0.6 &   2.5 &  12.0 &  12.4 &   2.4 &   7.6 &  16.2 \\
    & UM 2 &  16.3 &  18.2 &   5.7 &   3.5 &  13.9 &  16.0 &    0.3 &   0.5 &  12.8 &  10.4 &   1.2 &   6.7 &  14.6 \\
    & UM 3 &  21.9 &  18.5 &   8.3 &   3.9 &  17.5 &  17.7 &    0.2 &   0.2 &   9.1 &  13.2 &   1.6 &   7.6 &  16.4 \\
    & UM 4 &  17.0 &  23.7 &  10.3 &   7.0 &  18.0 &  17.6 &    3.4 &   4.7 &  18.3 &  11.9 &   3.5 &   8.8 &  17.9 \\
\cline{1-15}
\multirow{5}{*}{2B} & Nat &  38.6 &  29.7 &  19.6 &   6.6 &  45.2 &  49.9 &    5.5 &  12.5 &  48.7 &  40.5 &   7.5 &  14.9 &  45.1 \\
    & UM 1 &  30.9 &  26.5 &  19.3 &   7.6 &  38.8 &  33.0 &    3.2 &  11.4 &  35.1 &  31.8 &   3.8 &  16.9 &  28.0 \\
\bottomrule
\end{tabular}

\label{tab:bc-tccpp-p25}
\end{table*}

\begin{table*}[h]
\centering
\caption{\% changes in $pass@k$ compared to the models trained on the natural distribution for High Resource languages. For BC-HumanEval(HE), $k=100$. For BC-TP3(TP3), BC-Transcoder Python(TC-Py), and BC-Transcoder C++(TC-C++), $k=25$. The {\color{red} cells} represent the worst value for that language for that size and dataset. The {\color{green} cells} represent the best value for that language for that size and dataset.}
\begin{tabular}{lll|ccccccc|c}
\toprule
{DS} & {Size} & {Dist.} & {Java} & {Python} & {C++} & {PHP} & {TS} & {JS} & {Go} & {Mean} \\
\midrule
\multirow[c]{12}{*}{HE} & \multirow[c]{4}{*}{1B} & UM 1 & \cellcolor{red} \color{white} -29.0 & \cellcolor{red} \color{white} -15.0 & -30.3 & -12.4 & \cellcolor{red} \color{white} -20.3 & -4.1 & -19.6 & \cellcolor{red} \color{white} -18.7 \\
 &  & UM 2 & -19.6 & \cellcolor{green} -9.2 & \cellcolor{green} -27.0 & -13.4 & -14.7 & \cellcolor{green} 0.2 & \cellcolor{red} \color{white} -20.1 & -14.8 \\
 &  & UM 3 & \cellcolor{green} -12.1 & -9.7 & \cellcolor{red} \color{white} -41.7 & \cellcolor{red} \color{white} -14.6 & -16.7 & -3.3 & \cellcolor{green} 3.8 & \cellcolor{green} -13.5 \\
 &  & UM 4 & -18.9 & -11.9 & -28.2 & \cellcolor{green} -10.8 & \cellcolor{green} -12.1 & \cellcolor{red} \color{white} -5.1 & -17.2 & -14.9 \\
\cline{2-11}
 & \multirow[c]{4}{*}{2B} & UM 1 & \cellcolor{green} -15.2 & \cellcolor{red} \color{white} -20.3 & -15.9 & -17.1 & \cellcolor{green} -7.9 & -8.3 & \cellcolor{green} -5.6 & -12.9 \\
 &  & UM 2 & \cellcolor{green} -15.2 & -13.7 & -18.6 & -22.2 & \cellcolor{red} \color{white} -12.7 & \cellcolor{red} \color{white} -12.7 & \cellcolor{red} \color{white} -18.6 & \cellcolor{red} \color{white} -16.2 \\
 &  & UM 3 & -16.9 & -18.7 & \cellcolor{green} -12.7 & \cellcolor{green} -16.3 & -8.0 & \cellcolor{green} -7.3 & -6.7 & \cellcolor{green} -12.4 \\
\bottomrule
\end{tabular}
\label{tab:hr-pct-change}
\end{table*}

\begin{table*}
\centering
\caption{\% change of $pass@k$ compared to the models trained on the natural distribution for low resource languages languages. For BC-HumanEval(HE), $k=100$. For BC-TP3(TP3), BC-Transcoder Python(TC-Py), and BC-Transcoder C++(TC-C++), $k=25$. The {\color{red} cells} represent the worst value for that language for that size and dataset. The {\color{green} cells} represent the best value for that language for that size and dataset.}
\begin{tabular}{lll|ccccccc|c}
\toprule
{DS} & {Size} & {Dist.} & {Dart} & {Lua} & {Rust} & {C\#} & {R} & {Julia} & {HS} & {Mean} \\
\midrule
\multirow[c]{12}{*}{HE} & \multirow[c]{4}{*}{1B} & UM 1 & \cellcolor{green} -2.8 & 33.8 & \cellcolor{red} \color{white} 4.6 & \cellcolor{red} \color{white} 68.5 & \cellcolor{red} \color{white} 100.0 & \cellcolor{red} \color{white} 191.9 & \cellcolor{red} \color{white} 205.9 & \cellcolor{red} \color{white} 86.0 \\
 &  & UM 2 & -4.1 & 38.3 & \cellcolor{green} 19.0 & 98.1 & 161.7 & 254.7 & 238.7 & 115.2 \\
 &  & UM 3 & \cellcolor{red} \color{white} -6.4 & \cellcolor{red} \color{white} 30.8 & 15.5 & 87.1 & 162.7 & 218.4 & \cellcolor{green} 308.9 & 116.7 \\
 &  & UM 4 & -3.9 & \cellcolor{green} 46.5 & 14.8 & \cellcolor{green} 115.2 & \cellcolor{green} 166.9 & \cellcolor{green} 272.6 & 294.5 & \cellcolor{green} 129.5 \\
\cline{2-11}
 & \multirow[c]{4}{*}{2B} & UM 1 & 15.6 & -1.6 & \cellcolor{green} 12.7 & \cellcolor{red} \color{white} 59.4 & \cellcolor{red} \color{white} 28.6 & \cellcolor{red} \color{white} 145.2 & \cellcolor{red} \color{white} 147.7 & \cellcolor{red} \color{white} 58.2 \\
 &  & UM 2 & 21.6 & \cellcolor{red} \color{white} -5.9 & \cellcolor{red} \color{white} 3.4 & 71.2 & 44.9 & 172.9 & 161.5 & 67.1 \\
 &  & UM 3 & \cellcolor{red} \color{white} 12.2 & \cellcolor{green} 8.9 & 8.2 & \cellcolor{green} 78.6 & \cellcolor{green} 68.9 & 173.5 & \cellcolor{green} 177.9 & \cellcolor{green} 75.4 \\
\bottomrule
\end{tabular}
\label{tab:lr-pct-change}
\end{table*}

\newpage

\begin{table*}
\centering
\caption{Number of Questions passed for BC-HumanEval(HE) and TP3. BC-HE has 161 total problems and TP3 has 370 total problems. $S$ is the size of the model, and $D$ is the distribution it was trained on. P is the PaLM distribution while PC is the PaLM-Coder distribution. Languages are sorted from high to low resource. {\color[HTML]{5DB975} \textbf{Green}} values are the best values for that language, while {\color[HTML]{5DB975} \textbf{red}} values are the worst.}
\label{appendix:q-passed}
\begin{tabular}{lll|cccccccccccccc}
\toprule
$N$ & $S$ & $D$  & Java & Py & C++ & PHP & TS & JS & Go & Dart & Lua & Rust & C\# & R & Julia & HS \\
\midrule
\multirow[c]{19}{*}{HE} & \multirow[c]{5}{*}{1B} & N & \color[HTML]{5DB975} \bfseries 46 & \color[HTML]{5DB975} \bfseries 44 & \color[HTML]{5DB975} \bfseries 44 & \color[HTML]{5DB975} \bfseries 32 & \color[HTML]{5DB975} \bfseries 44 & \color[HTML]{5DB975} \bfseries 38 & \color[HTML]{5DB975} \bfseries 31 & \color[HTML]{5DB975} \bfseries 28 & \color[HTML]{FF847A} \bfseries 18 & 27 & \color[HTML]{FF847A} \bfseries 13 & \color[HTML]{FF847A} \bfseries 8 & \color[HTML]{FF847A} \bfseries 9 & \color[HTML]{FF847A} \bfseries 5 \\
 &  & U1 & \color[HTML]{FF847A} \bfseries 33 & \color[HTML]{FF847A} \bfseries 38 & 32 & 30 & \color[HTML]{FF847A} \bfseries 34 & 36 & 23 & 27 & 24 & \color[HTML]{FF847A} \bfseries 26 & 26 & 17 & 23 & 13 \\
 &  & U2 & 38 & 39 & 33 & \color[HTML]{FF847A} \bfseries 28 & 38 & \color[HTML]{5DB975} \bfseries 38 & \color[HTML]{FF847A} \bfseries 21 & 26 & 26 & \color[HTML]{5DB975} \bfseries 32 & 28 & \color[HTML]{5DB975} \bfseries 20 & 30 & 17 \\
 &  & U3 & 43 & 41 & \color[HTML]{FF847A} \bfseries 25 & 29 & 38 & 37 & \color[HTML]{5DB975} \bfseries 31 & \color[HTML]{FF847A} \bfseries 25 & 24 & 29 & 28 & \color[HTML]{5DB975} \bfseries 20 & 25 & \color[HTML]{5DB975} \bfseries 19 \\
 &  & U4 & 41 & 40 & 32 & 31 & 39 & \color[HTML]{FF847A} \bfseries 34 & 23 & 26 & \color[HTML]{5DB975} \bfseries 28 & \color[HTML]{5DB975} \bfseries 32 & \color[HTML]{5DB975} \bfseries 32 & 19 & \color[HTML]{5DB975} \bfseries 32 & 18 \\
\bottomrule
\end{tabular}
\end{table*}

\begin{table*}
\centering
\caption{Number of Questions passed for Transcoder. There are a total of 524 questions, and $N$ represents the source language. $S$ is the size of the model, and $D$ is the distribution it was trained on. P is the PaLM distribution while PC is the PaLM-Coder distribution. Languages are sorted from high to low resource. {\color[HTML]{5DB975} \textbf{Green}} values are the best values for that language, while {\color[HTML]{5DB975} \textbf{red}} values are the worst.}
\label{appendix:q-passed-tc}
\begin{tabular}{lll|cccccccccccccc}
\toprule
$N$ & $S$ & $D$  & Java & Py & C++ & PHP & TS & JS & Go & Dart & Lua & Rust & C\# & R & Julia & HS \\
\midrule
\multirow[c]{19}{*}{Py} & \multirow[c]{5}{*}{1B} & N & \color[HTML]{5DB975} \bfseries 118 &  & \color[HTML]{5DB975} \bfseries 124 & \color[HTML]{5DB975} \bfseries 62 & \color[HTML]{5DB975} \bfseries 96 & \color[HTML]{5DB975} \bfseries 103 & \color[HTML]{5DB975} \bfseries 70 & \color[HTML]{5DB975} \bfseries 41 & \color[HTML]{5DB975} \bfseries 52 & \color[HTML]{5DB975} \bfseries 68 & \color[HTML]{5DB975} \bfseries 99 & 20 & 24 & \color[HTML]{FF847A} \bfseries 15 \\
 &  & U1 & \color[HTML]{FF847A} \bfseries 41 &  & \color[HTML]{FF847A} \bfseries 13 & \color[HTML]{FF847A} \bfseries 11 & \color[HTML]{FF847A} \bfseries 51 & \color[HTML]{FF847A} \bfseries 40 & 41 & 10 & 10 & \color[HTML]{FF847A} \bfseries 43 & \color[HTML]{FF847A} \bfseries 28 & \color[HTML]{FF847A} \bfseries 14 & \color[HTML]{FF847A} \bfseries 9 & 25 \\
 &  & U2 & 62 &  & 25 & 25 & 73 & 78 & 28 & 19 & \color[HTML]{FF847A} \bfseries 4 & 46 & 33 & 17 & \color[HTML]{5DB975} \bfseries 27 & 26 \\
 &  & U3 & 51 &  & 43 & \color[HTML]{FF847A} \bfseries 11 & 52 & 66 & 17 & \color[HTML]{FF847A} \bfseries 8 & 9 & \color[HTML]{FF847A} \bfseries 43 & 46 & 19 & 17 & 30 \\
 &  & U4 & 58 &  & 34 & 43 & 70 & 81 & \color[HTML]{FF847A} \bfseries 8 & 21 & 12 & 51 & 40 & \color[HTML]{5DB975} \bfseries 25 & 22 & \color[HTML]{5DB975} \bfseries 42 \\
\Xcline{2-17}{0.5pt}
 & \multirow[c]{5}{*}{2B} & N & 191 &  & 225 & \color[HTML]{5DB975} \bfseries 197 & 160 & 231 & 114 & 76 & 47 & \color[HTML]{FF847A} \bfseries 78 & 140 & \color[HTML]{5DB975} \bfseries 46 & 36 & 41 \\
 &  & U1 & 182 &  & 154 & 115 & \color[HTML]{FF847A} \bfseries 80 & \color[HTML]{FF847A} \bfseries 160 & 89 & 78 & 43 & 102 & 144 & 41 & 22 & \color[HTML]{FF847A} \bfseries 36 \\
\bottomrule
\end{tabular}
\end{table*}

\begin{table*}
\centering
\caption{Metrics for HR languages on BC-HumanEval for all models. $\Delta$ is the mean change of each of the displayed langauges when compared to the natural.\% Failed tests is the percent of predictions that did not have any errors, but failed a test. \% Error is the percent of predictions that had either a runtime or compilation error. \% Timed Out is the percent of predictions that timed out. The time out was set to 10 for all languages except for Java and TS, which was 15. \% Passed is the percent of predictions that passed all test cases. \% Passed One is the percent of predictions that passed at least one test case, but failed.\% Tests Passed is the mean percent of test cases passed per problem for all predictions.}
\label{appendix:he-qual-hr}
\begin{tabular}{ll|ccccccc|c}
\toprule
Metric & $D$  & Java & Py & C++ & PHP & TS & JS & Go & $\Delta$\\
\midrule
\multirow[c]{3}{*}{\% Error} & N & 25.32 & 19.36 & 17.80 & 8.61 & 21.53 & 11.66 & 49.02 &  \\
 & $U$1 & 28.85 & 17.45 & 19.83 & 10.25 & 21.53 & 11.00 & 47.23 & 0.40 \\
 & $U$2 & 34.08 & 18.16 & 19.80 & 8.65 & 20.87 & 9.67 & 50.13 & 1.15 \\
\Xhline{1pt}
\multirow[c]{3}{*}{\% Failed Test} & N & 59.94 & 65.12 & 64.94 & 78.45 & 63.92 & 74.60 & 42.02 &  \\
 & $U$1 & 58.12 & 71.33 & 63.03 & 79.32 & 64.41 & 76.44 & 44.85 & 1.22 \\
 & $U$2 & 54.34 & 68.89 & 66.97 & 80.25 & 66.59 & 77.56 & 42.34 & 1.13 \\
\Xhline{1pt}
\multirow[c]{3}{*}{\% Passed} & N & 13.45 & 14.60 & 12.70 & 10.12 & 11.71 & 12.29 & 8.15 &  \\
 & $U$1 & 11.57 & 10.68 & 11.29 & 8.41 & 11.69 & 11.64 & 7.50 & -1.46 \\
 & $U$2 & 10.16 & 11.93 & 11.05 & 8.37 & 11.29 & 11.30 & 6.96 & -1.71 \\
\Xhline{1pt}
\multirow[c]{3}{*}{\% Passed One} & N & 47.26 & 46.20 & 43.77 & 46.70 & 45.32 & 49.87 & 28.82 &  \\
 & $U$1 & 44.68 & 42.83 & 42.80 & 43.60 & 45.95 & 48.47 & 30.39 & -1.32 \\
 & $U$2 & 41.69 & 43.92 & 43.38 & 43.02 & 46.13 & 47.87 & 28.69 & -1.89 \\
\Xhline{1pt}
\multirow[c]{3}{*}{\% Tests Passed} & N & 33.46 & 33.77 & 31.07 & 28.84 & 30.71 & 32.78 & 20.03 &  \\
 & $U$1 & 30.45 & 28.69 & 29.25 & 26.29 & 31.42 & 31.78 & 20.21 & -1.79 \\
 & $U$2 & 27.44 & 29.58 & 28.64 & 25.49 & 30.44 & 30.61 & 18.75 & -2.81 \\
\Xhline{1pt}
\multirow[c]{3}{*}{\% Timed Out} & N & 1.29 & 0.93 & 4.57 & 2.82 & 2.84 & 1.45 & 0.80 &  \\
 & $U$1 & 1.45 & 0.54 & 5.86 & 2.02 & 2.37 & 0.92 & 0.42 & -0.16 \\
 & $U$2 & 1.42 & 1.02 & 2.18 & 2.74 & 1.25 & 1.47 & 0.57 & -0.58 \\
\bottomrule
\end{tabular}
\end{table*}

\begin{table*}
\centering
\caption{Metrics for LR languages on BC-HumanEval for all models. $\Delta$ is the mean change of each of the displayed langauges when compared to the natural.\% Failed tests is the percent of predictions that did not have any errors, but failed a test. \% Error is the percent of predictions that had either a runtime or compilation error. \% Timed Out is the percent of predictions that timed out. The time out was set to 10 for all languages except for Java and TS, which was 15. \% Passed is the percent of predictions that passed all test cases. \% Passed One is the percent of predictions that passed at least one test case, but failed.\% Tests Passed is the mean percent of test cases passed per problem for all predictions.}
\label{appendix:he-qual-lr}
\begin{tabular}{ll|ccccccc|c}
\toprule
Metric & $D$  & Dart & Lua & Rust & C\# & R & Julia & HS & $\Delta$\\
\midrule
\multirow[c]{3}{*}{\% Error} & N & 62.06 & 31.31 & 51.61 & 43.80 & 70.08 & 68.90 & 85.70 &  \\
 & $U$1 & 56.05 & 23.39 & 48.20 & 44.40 & 54.24 & 50.51 & 70.80 & -9.41 \\
 & $U$2 & 54.64 & 20.28 & 42.62 & 41.11 & 52.07 & 47.10 & 69.75 & -12.27 \\
\Xhline{1pt}
\multirow[c]{3}{*}{\% Failed Test} & N & 28.71 & 57.98 & 38.51 & 45.42 & 26.66 & 25.72 & 11.67 &  \\
 & $U$1 & 34.37 & 66.01 & 41.50 & 46.84 & 42.05 & 42.28 & 24.69 & 9.01 \\
 & $U$2 & 35.26 & 69.21 & 45.62 & 48.56 & 43.26 & 44.92 & 25.52 & 11.10 \\
\Xhline{1pt}
\multirow[c]{3}{*}{\% Passed} & N & 8.74 & 8.60 & 8.74 & 9.94 & 2.99 & 4.75 & 1.81 &  \\
 & $U$1 & 9.19 & 9.23 & 9.47 & 7.97 & 3.46 & 6.57 & 3.08 & 0.49 \\
 & $U$2 & 9.27 & 8.74 & 10.73 & 8.86 & 3.98 & 6.83 & 3.57 & 0.92 \\
\Xhline{1pt}
\multirow[c]{3}{*}{\% Passed One} & N & 25.51 & 40.39 & 28.48 & 36.26 & 15.62 & 25.07 & 8.29 &  \\
 & $U$1 & 30.95 & 42.98 & 30.94 & 36.57 & 23.29 & 34.48 & 16.90 & 5.21 \\
 & $U$2 & 31.58 & 35.78 & 33.45 & 36.41 & 26.42 & 33.11 & 17.55 & 4.95 \\
\Xhline{1pt}
\multirow[c]{3}{*}{\% Tests Passed} & N & 19.43 & 24.31 & 20.53 & 25.49 & 8.70 & 13.79 & 5.13 &  \\
 & $U$1 & 22.31 & 26.00 & 22.31 & 23.45 & 12.03 & 19.59 & 9.53 & 2.55 \\
 & $U$2 & 22.32 & 22.79 & 24.36 & 23.92 & 13.68 & 19.25 & 10.48 & 2.77 \\
\Xhline{1pt}
\multirow[c]{3}{*}{\% Timed Out} & N & 0.49 & 2.10 & 1.13 & 0.85 & 0.28 & 0.63 & 0.82 &  \\
 & $U$1 & 0.39 & 1.38 & 0.82 & 0.80 & 0.25 & 0.64 & 1.43 & -0.09 \\
 & $U$2 & 0.83 & 1.78 & 1.04 & 1.46 & 0.69 & 1.14 & 1.16 & 0.26 \\
\bottomrule
\end{tabular}
\end{table*}

\begin{table*}
\centering
\caption{Metrics for HR languages on TP3 for all models. $\Delta$ is the mean change of each of the displayed langauges when compared to the natural.\% Failed tests is the percent of predictions that did not have any errors, but failed a test. \% Error is the percent of predictions that had either a runtime or compilation error. \% Timed Out is the percent of predictions that timed out. The time out was set to 10 for all languages except for Java and TS, which was 15. \% Passed is the percent of predictions that passed all test cases. \% Passed One is the percent of predictions that passed at least one test case, but failed.\% Tests Passed is the mean percent of test cases passed per problem for all predictions.}
\label{appendix:tp3-qual-hr}
\begin{tabular}{ll|cccccc|c}
\toprule
Metric & $D$  & Java & C++ & PHP & TS & JS & Go & $\Delta$\\
\midrule
\multirow[c]{3}{*}{\% Error} & N & 60.94 & 49.05 & 59.66 & 62.13 & 60.44 & 92.53 &  \\
 & $U$1 & 65.04 & 56.15 & 58.08 & 52.67 & 47.27 & 86.54 & -3.17 \\
 & $U$2 & 52.71 & 31.20 & 50.03 & 56.74 & 47.73 & 82.85 & -10.58 \\
\Xhline{1pt}
\multirow[c]{3}{*}{\% Failed Test} & N & 25.09 & 16.37 & 29.14 & 12.31 & 28.45 & 3.54 &  \\
 & $U$1 & 23.17 & 17.19 & 32.55 & 19.78 & 38.94 & 7.71 & 4.07 \\
 & $U$2 & 32.95 & 19.17 & 37.92 & 17.92 & 39.45 & 12.63 & 7.52 \\
\Xhline{1pt}
\multirow[c]{3}{*}{\% Passed} & N & 9.40 & 6.54 & 10.39 & 7.33 & 10.91 & 3.91 &  \\
 & $U$1 & 7.67 & 6.10 & 8.62 & 9.56 & 13.52 & 5.66 & 0.44 \\
 & $U$2 & 8.30 & 4.12 & 9.80 & 7.73 & 11.68 & 4.36 & -0.42 \\
\Xhline{1pt}
\multirow[c]{3}{*}{\% Passed One} & N & 28.57 & 15.97 & 27.25 & 12.20 & 31.76 & 3.77 &  \\
 & $U$1 & 27.17 & 16.99 & 29.46 & 19.59 & 43.19 & 8.45 & 4.22 \\
 & $U$2 & 37.95 & 17.98 & 34.96 & 17.14 & 40.92 & 12.97 & 7.07 \\
\Xhline{1pt}
\multirow[c]{3}{*}{\% Tests Passed} & N & 23.31 & 14.69 & 24.17 & 13.66 & 26.44 & 5.81 &  \\
 & $U$1 & 20.82 & 14.76 & 23.62 & 19.67 & 34.89 & 9.80 & 2.58 \\
 & $U$2 & 26.81 & 13.21 & 27.44 & 16.55 & 31.70 & 10.72 & 3.06 \\
\Xhline{1pt}
\multirow[c]{3}{*}{\% Timed Out} & N & 4.57 & 28.03 & 0.81 & 18.23 & 0.20 & 0.02 &  \\
 & $U$1 & 4.13 & 20.56 & 0.76 & 17.98 & 0.28 & 0.09 & -1.34 \\
 & $U$2 & 6.04 & 45.51 & 2.24 & 17.61 & 1.14 & 0.16 & 3.47 \\
\bottomrule
\end{tabular}
\end{table*}

\begin{table*}
\centering
\caption{Metrics for LR languages on TP3 for all models. $\Delta$ is the mean change of each of the displayed langauges when compared to the natural.\% Failed tests is the percent of predictions that did not have any errors, but failed a test. \% Error is the percent of predictions that had either a runtime or compilation error. \% Timed Out is the percent of predictions that timed out. The time out was set to 10 for all languages except for Java and TS, which was 15. \% Passed is the percent of predictions that passed all test cases. \% Passed One is the percent of predictions that passed at least one test case, but failed.\% Tests Passed is the mean percent of test cases passed per problem for all predictions.}
\label{appendix:tp3-qual-lr}
\begin{tabular}{ll|ccccccc|c}
\toprule
Metric & $D$  & Dart & Lua & Rust & C\# & R & Julia & HS & $\Delta$\\
\midrule
\multirow[c]{3}{*}{\% Error} & N & 90.12 & 93.45 & 84.47 & 80.85 & 97.34 & 89.43 & 89.60 &  \\
 & $U$1 & 79.83 & 85.22 & 77.93 & 80.02 & 95.20 & 83.65 & 89.39 & -4.86 \\
 & $U$2 & 80.96 & 86.36 & 72.00 & 71.00 & 92.17 & 81.19 & 84.96 & -8.09 \\
\Xhline{1pt}
\multirow[c]{3}{*}{\% Failed Test} & N & 4.78 & 5.42 & 11.54 & 13.10 & 2.00 & 4.23 & 7.96 &  \\
 & $U$1 & 12.24 & 10.25 & 16.07 & 14.06 & 3.60 & 6.73 & 7.98 & 3.13 \\
 & $U$2 & 12.73 & 9.82 & 21.21 & 21.30 & 6.38 & 9.20 & 11.34 & 6.13 \\
\Xhline{1pt}
\multirow[c]{3}{*}{\% Passed} & N & 5.07 & 0.94 & 3.77 & 5.87 & 0.62 & 3.51 & 1.31 &  \\
 & $U$1 & 7.83 & 4.22 & 5.84 & 5.76 & 1.20 & 5.92 & 1.74 & 1.63 \\
 & $U$2 & 6.09 & 3.11 & 6.18 & 7.14 & 1.32 & 6.10 & 2.75 & 1.65 \\
\Xhline{1pt}
\multirow[c]{3}{*}{\% Passed One} & N & 5.28 & 5.51 & 11.77 & 14.87 & 0.95 & 7.15 & 7.83 &  \\
 & $U$1 & 13.96 & 11.43 & 16.51 & 16.31 & 1.83 & 11.11 & 7.97 & 3.68 \\
 & $U$2 & 14.57 & 9.88 & 22.00 & 24.70 & 5.34 & 14.85 & 11.45 & 7.06 \\
\Xhline{1pt}
\multirow[c]{3}{*}{\% Tests Passed} & N & 7.76 & 3.55 & 9.59 & 13.23 & 1.10 & 6.87 & 5.18 &  \\
 & $U$1 & 14.74 & 9.62 & 14.10 & 13.74 & 2.11 & 11.03 & 5.77 & 3.40 \\
 & $U$2 & 13.33 & 7.78 & 17.00 & 19.01 & 3.92 & 12.77 & 8.40 & 4.99 \\
\Xhline{1pt}
\multirow[c]{3}{*}{\% Timed Out} & N & 0.02 & 0.20 & 0.22 & 0.18 & 0.03 & 2.82 & 1.12 &  \\
 & $U$1 & 0.11 & 0.31 & 0.16 & 0.16 & 0.01 & 3.70 & 0.89 & 0.11 \\
 & $U$2 & 0.22 & 0.71 & 0.60 & 0.56 & 0.13 & 3.51 & 0.96 & 0.30 \\
\bottomrule
\end{tabular}
\end{table*}

\end{document}